\documentclass[11pt]{article}

\usepackage[final]{acl}

\usepackage{times}
\usepackage{latexsym}

\usepackage[T1]{fontenc}

\usepackage[utf8]{inputenc}

\usepackage{microtype}

\usepackage{inconsolata}

\usepackage{graphicx}
\usepackage{amsmath}
\usepackage{multirow}
\usepackage{pifont}
\usepackage{makecell}
\usepackage{subcaption}
\usepackage{colortbl} 
\usepackage{amssymb}
\usepackage{lineno}
\usepackage{soul}
\usepackage{enumitem}

\newcommand{\methodname}[1]{\textcolor{blue}{\texttt{FORGE}}}
\newcommand{\papertitle}[1]{\methodname{}: Forensic Reasoning with Grounded Evidence}

\newcommand{\red}[1]{red!25}
\newcommand{\blue}[1]{blue!15}

\title{\papertitle{}\textsuperscript{*}}

\author{
    Rohit Kundu$^{1,2}$, Shan Jia$^2$, Vishal Mohanty$^2$, Athula Balachandran$^2$, Amit K. Roy-Chowdhury$^1$\\
    \textsuperscript{\rm 1}University of California, Riverside; 
    \textsuperscript{\rm 2}Google LLC\\
    {\tt\small \{rohit.kundu@email, amitrc@ece\}.ucr.edu; \{rohitkun, shanjia, vishalmohanty, athula\}@google.com}
}

\begin{document}

\twocolumn[{%
\renewcommand\twocolumn[1][]{#1}%
\maketitle
\begin{center}
    \centering
    \includegraphics[width=\textwidth]{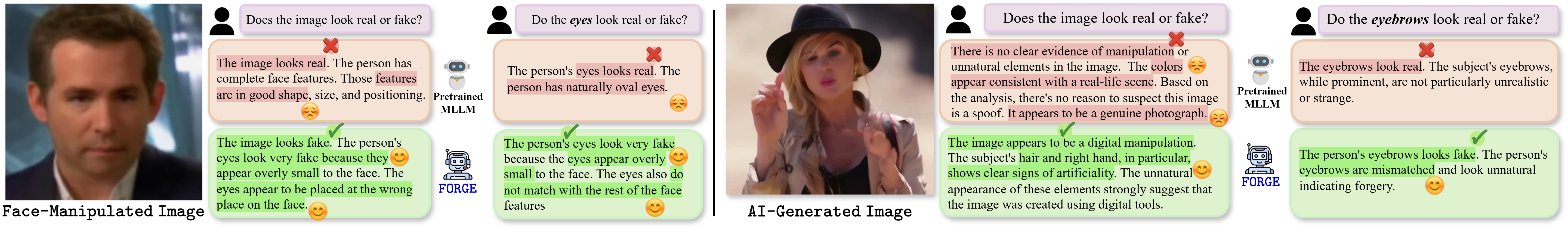}
    \captionof{figure}{\textbf{Illustration of Examples}: Comparison of pretrained PaliGemma2 \cite{steiner2024paligemma} multimodal large language model's (MLLM) DeepFake reasoning against our proposed \methodname{} framework on a face-manipulated image (left) and a fully AI-generated image (right). The pretrained PaliGemma2 fails to detect subtle inconsistencies in the manipulated images, resulting in incorrect explanations. In contrast, \methodname{} provides accurate predictions along with detailed, fine-grained reasoning for the detected manipulations.}
    \label{fig:teaser}
\end{center}%
}]
\begingroup
\renewcommand{\thefootnote}{*}
\footnotetext{This paper has been accepted at Findings of the Association for Computational Linguistics: EMNLP 2026}
\endgroup

\begin{abstract}
    Forensic deepfake analysis demands more than binary classification: investigators need region-grounded natural language explanations they can verify against the image. Multimodal large language models (MLLMs) are a natural fit, but pretrained MLLMs fail systematically, producing globally coherent text that misses the small localized cues defining manipulations. We argue this is an inductive bias problem rather than a capacity issue: the image-text contrastive objective training MLLM visual encoders optimizes for whole-image semantic summaries, not patch-level forensic detail. The same mismatch explains why prior deepfake reasoning methods target either face manipulation or fully AI-generated content, never both.

We propose \methodname{}, which addresses the mismatch by routing a second visual stream into the language model from a Vision-Only Model (VOM) trained on dense patch prediction rather than image-text alignment. The MLLM's native encoder and the VOM operate on a shared patch grid, which lets us interleave their tokens with preserved spatial correspondence; we show this beats naive concatenation. A two-stage adapter training protocol (generic image-caption alignment, then joint task-specific optimization) prevents the localized stream from overfitting to training-domain manipulations. Across face-manipulated and fully synthetic content, \methodname{} produces region-referential explanations answering fine-grained attribute queries (``Does the \texttt{eyes}/\texttt{nose}/\texttt{mouth} look real or fake?'') and substantially outperforms in-domain baselines on cross-domain evaluations; region-specific evaluation and human studies confirm explanation faithfulness.
\end{abstract}

\section{Introduction}\label{sec:intro}
Generative models for images and video have become both more accessible and more capable \cite{fluxgenerator, runwayresearch, wang2023modelscope, opensora}, producing manipulated content that humans struggle to identify \cite{li2019faceshifter, thies2019deferred}. The forensic response to this has been a long line of detection methods that classify content as real or fake \cite{cheng2024can, kundu2024towards}. But in deployment settings, a binary label is rarely sufficient. An investigator presented with a ``fake'' verdict needs to know which regions of the image triggered the decision and what about those regions looks manipulated, in natural language they can verify against the image itself. Naturally, we ask: \textit{can multimodal large language models (MLLMs), optimized to describe images in broad strokes, be redirected to identify the small localized cues that define forensic manipulations?}

Recent work has explored using MLLMs for DeepFake reasoning, but the literature has fractured along the content type axis. One body of work targets face manipulation, where a real face image is altered by face swap, expression transfer, or attribute editing \cite{zhang2024common, chen2024x2dfd, sun2025towards, guo2025rethinking}. A separate body targets fully AI-generated content from text-to-image, text-to-video, or image-to-video models \cite{song2024learning, yang2025heie}. The two lines have proceeded essentially independently, with different datasets, different architectural choices, and different evaluation protocols. To our knowledge, no prior MLLM-based DeepFake reasoning method jointly addresses both content categories within a single framework.

This bifurcation is not coincidental. Pretrained MLLMs, evaluated on either task, fail in a specific and consistent way: they produce globally fluent descriptions that miss the small localized cues that distinguish manipulated from authentic content (Figure~\ref{fig:teaser}). We argue this is not a capacity issue, since using larger MLLMs for the problem does not resolve it \cite{jia2024can}, but an inductive bias issue. The image-text contrastive objective used to train MLLM visual encoders (CLIP \cite{radford2021learning}, SigLIP \cite{zhai2023sigmoid}, and variants) optimizes representations for whole-image semantic summarization, the wrong target for forensic reasoning where the evidence lives in localized patches \cite{dahou2025vision, tong2024eyes}. Naively fine-tuning the MLLM on forensic data papers over this on in-domain test sets but does not transfer across content categories or across manipulation types within a category.

\begin{figure}
    \centering
    \includegraphics[width=\linewidth]{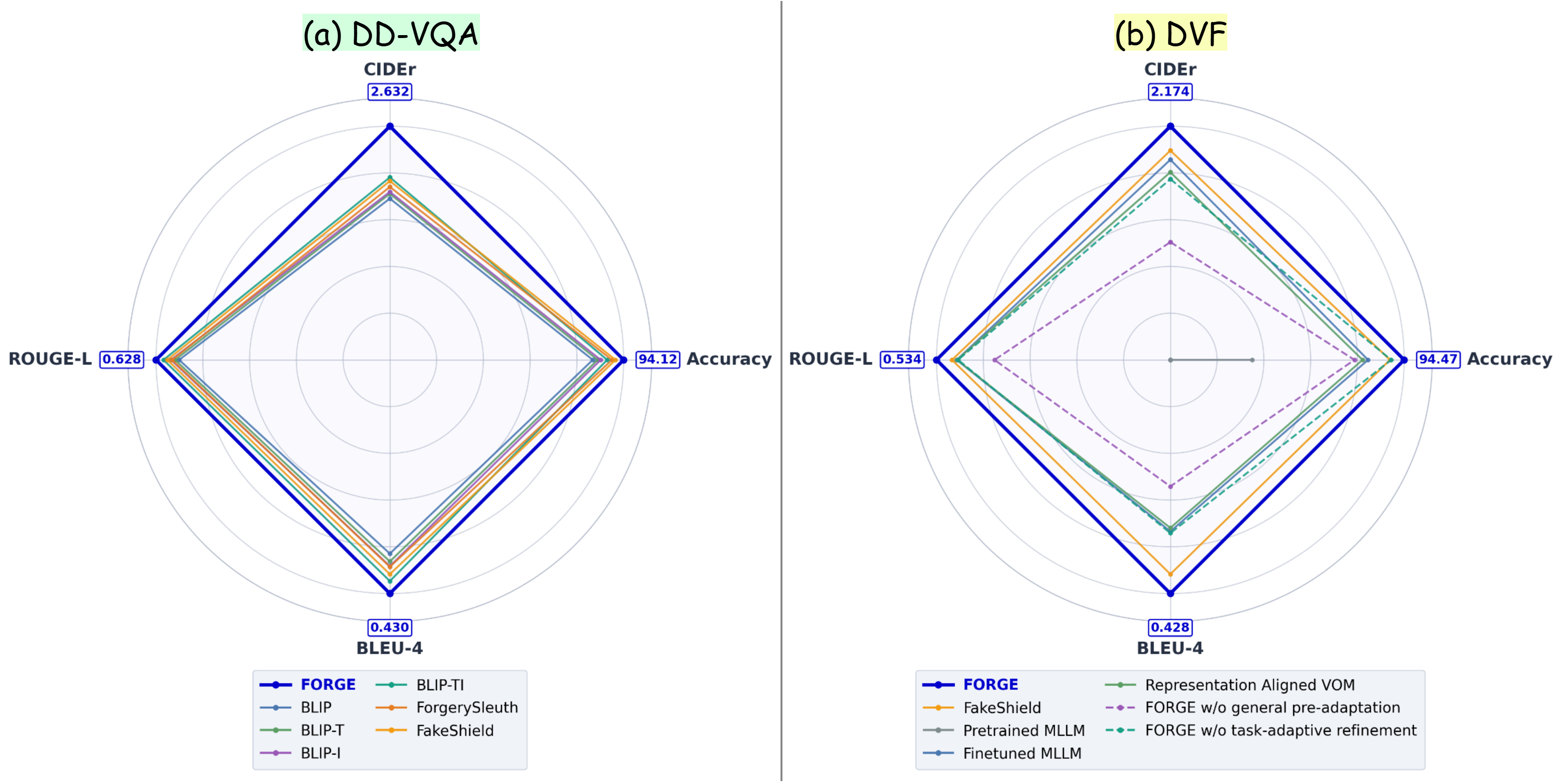}
    \caption{Comparison of \methodname{} and state-of-the-art DeepFake reasoning methods on two benchmarks. (a) DD-VQA \cite{zhang2024common} (face-manipulated) and (b) DVF \cite{song2024learning} (fully AI-generated): Axis markings indicate the actual scores for Accuracy and scaled scores for BLEU-4, and ROUGE-L, and CIDEr. \methodname{} consistently outperforms competitors across all explanation and detection metrics, despite their specialization for face manipulation in (a) and significantly outperforms pretrained and simple finetuned MLLMs in (b), showing that our strategic grounding is essential for reasoning.}
    \label{fig:teaser_results}
\end{figure}

We propose \methodname{}, which addresses the inductive bias mismatch directly. Rather than relying on the MLLM's contrastive encoder alone, we route a second visual stream into the language model from a Vision-Only Model (VOM) trained on dense patch prediction rather than image-text alignment, an encoder whose features are inherently localized by construction. The two streams are fused through a token-interleaved combination at a shared patch grid, which preserves spatial correspondence in a way that naive concatenation does not (Table~\ref{tab:ablation_features}). A lightweight adapter aligns the VOM stream to the LLM's embedding space, trained in two stages: first on generic image-caption data to establish general visual-linguistic alignment, then jointly with the LLM on forensic data. The first stage prevents the adapter from overfitting to training-domain manipulation patterns; ablating it collapses cross-domain performance (Table~\ref{tab:adapter_ablations}). Because the VOM stream operates at patch resolution and is decoupled from face-cropping preprocessing, the same framework handles face-manipulation cues (asymmetric eyes, blending artifacts) and fully-synthetic cues (implausible textures, scene-level inconsistency) without separate architectures or training pipelines. 

The contributions of this work are as follows.
\begin{itemize}[noitemsep,topsep=0pt,parsep=0pt,partopsep=0pt]
    \item We diagnose a systematic failure mode of pretrained MLLMs on DeepFake reasoning and trace it to the image-text contrastive training of their visual encoders, an inductive bias problem rather than a capacity problem, not addressable by scale or naive fine-tuning alone.

    
    \item We propose a dual-stream visual pathway that pairs the MLLM with a Vision-Only Model (VOM) trained on dense patch prediction, using token-interleaved fusion at a shared $14{\times}14$ patch resolution to preserve spatial correspondence; we show that this spatial alignment, not the combination of encoders alone, is what enables region-referential generation.

    
    \item We show that two-stage adapter training (general visual-linguistic alignment, followed by joint task-specific optimization) is necessary to preserve cross-domain generalization, ablating each stage to demonstrate the specific failure modes it prevents.

    
    \item \methodname{} is the first MLLM-based DeepFake reasoning framework to address both face-manipulated and fully synthetic content within a single architecture. We validate it through region-specific evaluation (eyes, nose, mouth for face manipulation; lighting, shadows, textures for fully synthetic content), human studies of explanation faithfulness, and cross-domain detection accuracy that outperforms methods trained in-domain on the evaluation datasets.
\end{itemize}

\begin{figure*}
    \centering
    \includegraphics[width=\textwidth]{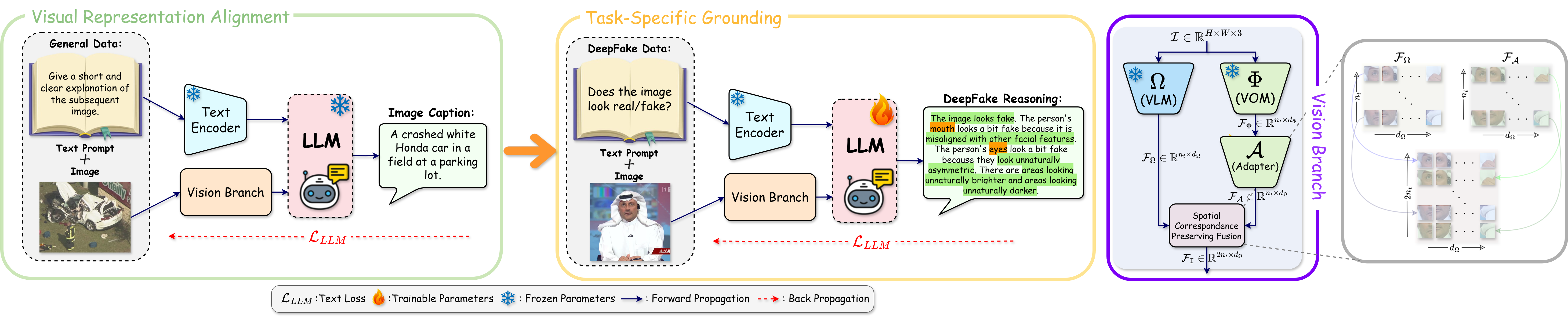}
    \caption{\textbf{Architecture of \methodname{}:} The framework routes two visual streams with complementary inductive biases into the language model. The contrastive encoder $\Omega$ (VLM) supplies global semantic features that are natively compatible with the LLM, while the vision-only model $\Phi$ (VOM) supplies dense patch features through a learnable adapter $\mathcal{A}$ that maps them into the LLM's embedding space. The two streams operate on a shared patch grid, which lets us combine them via spatial-correspondence-preserving fusion (right) without learned fusion layers. Training proceeds in two stages: visual representation alignment (left), where $\mathcal{A}$ is calibrated on general image-caption data with the LLM frozen, then task-specific grounding (center), where $\mathcal{A}$ and the LLM are jointly optimized on language-annotated deepfake data. The same architecture handles face-manipulated and fully AI-generated content.}
    \label{fig:overall}
\end{figure*}
\section{Related Work}

\noindent
\ul{\textbf{Detection without Reasoning:}} Early DeepFake detection methods \cite{li2018exposing, yang2019exposing} struggled with cross-dataset generalization, a problem subsequently addressed through identity-disentangled features \cite{dong2023implicit}, frequency-domain analysis \cite{corvi2023intriguing, corvi2023detection}, and reconstruction-based scores \cite{wang2023dire}. UNITE \cite{kundu2024towards} is the only method explicitly designed for both face-manipulated and fully synthetic content, using a transformer with attention-diversity loss, but like its predecessors it produces only binary labels. A separate thread surfaces detector decisions through visual saliency: Grad-CAM analyses \cite{tantaru2024weakly, xu2022supervised} highlight regions of high gradient response, and identity-prototype methods \cite{aghasanli2023interpretable} localize manipulations through distances to known identities. Both are limited as forensic tools: heatmaps are spatially coarse and show where the model attended rather than what is wrong, and prototype methods generalize poorly outside training identities. None of these approaches produce the natural language explanations one can verify against the image.

\noindent
\ul{\textbf{Face Manipulation Reasoning:}} One body of work targets face-manipulation DeepFakes (face swap, expression transfer, attribute editing) and generates natural language explanations through finetuned MLLMs. Early attempts evaluated GPT-4V \cite{achiam2023gpt} and Gemini-1.0 \cite{team2023gemini} in zero-shot settings \cite{jia2024can, shi2024shield}, finding that off-the-shelf MLLMs underperform task-specialized detectors. Subsequent methods finetune on annotated DeepFake data: DD-VQA \cite{zhang2024common} finetunes BLIP \cite{li2022blip}; X${}^2$-DFD \cite{chen2024x2dfd} finetunes LLaVA \cite{liu2024visual}; VLFFD \cite{sun2025towards} uses visual-language prompts to guide detection; M2F2-Det \cite{guo2025rethinking} combines CLIP with a Vicuna-7b decoder; ForgerySleuth \cite{sun2025forgerysleuth} adds a trace encoder for MLLM-based manipulation analysis; FFAA \cite{huang2024ffaa} proposes a multi-answer decision system for open-world face forgery analysis; and Fake-in-Facext \cite{qin2025fake} introduces a fine-grained facial concept tree for attribute-level explanation. These methods share two limitations relevant to our setting: they detect and crop faces as a preprocessing step (assuming a front-view face is always visible), and they are evaluated only on face-manipulated content and cannot describe the scene-level cues that define fully synthetic images.

\noindent
\ul{\textbf{Synthetic Content Reasoning:}} A symmetric body of work targets fully AI-generated content from text-to-image (T2I), text-to-video (T2V), and image-to-video (I2V) models, where manipulations affect entire scenes rather than localized regions. MM-Det \cite{song2024learning} finetunes LLaVA \cite{liu2024visual} on diffusion-generated video with text annotations to generate explanations. HEIE \cite{yang2025heie} produces hierarchical implausibility heatmaps with occasional textual descriptions, committing primarily to spatial output. More recent work explores preference optimization with auxiliary visual experts for AI-generated image detection \cite{zhou2025aigi} and two-stage scan-and-zoom strategies that locate suspicious regions before generating verdicts \cite{ji2025zoom}. Symmetrically to the face-manipulation methods, these systems are evaluated only on fully synthetic content and do not detect or explain face-manipulation cues that affect a small region of an otherwise authentic image.

\noindent
\underline{\textbf{Positioning of \methodname{}:}} The two bodies of work above have proceeded essentially independently, training on different datasets, using different architectures, and reporting different evaluation protocols. To the best of our knowledge, no prior MLLM-based DeepFake reasoning method addresses both content categories within a single framework. We argue in Section~\ref{sec:intro} that this bifurcation reflects the inductive bias mismatch of MLLM visual encoders rather than coincidence. Two recent methods come close to a unified treatment and warrant explicit positioning. FakeShield \cite{xu2025fakeshield} covers Photoshop tampering, DeepFake attribute editing, and partial AIGC inpainting (a real image with a region redrawn by Stable Diffusion), but does not address fully synthetic content where the entire image is generated from scratch (T2I, T2V, I2V). IVY-FAKE \cite{jiang2025ivy} unifies image and video AI-generated content detection with natural language explanations, but is not evaluated on face-manipulation benchmarks where a real face is altered. \methodname{} addresses the original bifurcation directly: face-manipulation cues that affect small regions of otherwise authentic images, and synthesis cues that affect entire scenes, within a single architecture trained on a single set of language-annotated data.
\section{Proposed Method}\label{sec:method}
\subsection{Problem Setup}\label{sec:problem_setup}
Unlike existing methods \cite{chen2024x2dfd, zhang2024common, sun2025towards, guo2025rethinking} that focus solely on face-manipulated DeepFakes, our model provides a unified framework for explainability in both face-manipulated and fully AI-generated images. Given an image $\mathcal{I} \in \mathbb{R}^{H \times W \times 3}$ and a question $\mathcal{Q}$ related to DeepFake detection, the objective of our model is to generate a detailed answer (ground truth answer: $\mathcal{E}$) that determines whether the image is real or manipulated and provides an explanation for the decision. Our goal is to find the optimal parameters $\theta$ of our model $f_\theta$ that minimize the following objective function:
\begin{equation}
    \underset{\theta}{\arg\min} \sum_{(\mathcal{I}, \mathcal{Q}, \mathcal{E}) \in \mathcal{D}} \mathcal{L}(f_\theta(\mathcal{I}, \mathcal{Q}), \mathcal{E})
\end{equation}
\noindent
where $\mathcal{D}$ is our training dataset of image-question-explanation triplets, and $\mathcal{L}$ is a loss function measuring discrepancy between generated and ground truth explanations. This formulation enables our model to learn accurate, detailed explanations for DeepFake images.

\subsection{Complementary Visual Inductive Biases}\label{subsec:foundation_models}
As argued in Sec.~\ref{sec:intro}, the failure mode of pretrained MLLMs on this task is an inductive bias mismatch: a visual encoder optimized for whole-image alignment with captions cannot reliably surface the patch-level cues that define face manipulations. We respond architecturally rather than by scaling. We leave the MLLM's native visual stream untouched and route a second stream into the language model from a VOM trained on dense patch prediction. The two streams are chosen for their structurally different training objectives, not as redundant feature extractors.

\noindent
\ul{\textbf{Global Context:}} The MLLM's native visual encoder ($\Omega$) is trained with image-text contrastive loss, which optimizes representations for whole-image alignment with caption embeddings. This objective is well-suited to scene-level reasoning, the level at which fully synthetic content typically reveals itself: composition, lighting, object arrangement, and semantic plausibility. $\Omega$ partitions $\mathcal{I}$ into non-overlapping $p\times p$ patches and produces,
\begin{equation}\label{eq:global}
\mathcal{F}_{\Omega} = \Omega(\mathcal{I}^{H\times W\times 3}) \in \mathbb{R}^{n_t\times d_{\Omega}},
\end{equation}
\noindent
where $n_t = \frac{H\times W}{p^2}$. These $n_t$ tokens carry the holistic scene signal that the LLM already consumes natively, so we route them into the language model directly without transformation.

\begin{figure}
    \centering
    \includegraphics[width=\columnwidth]{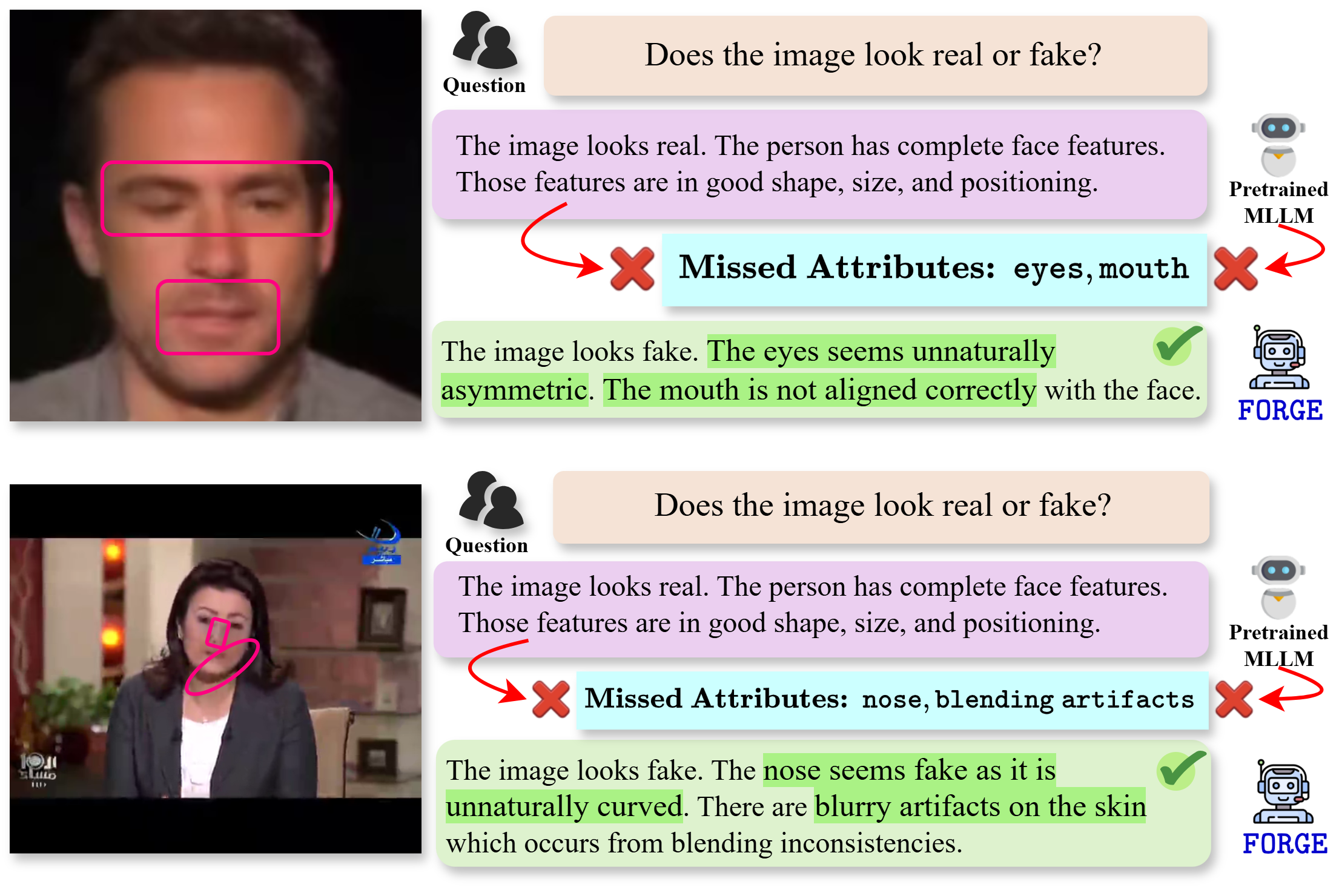}
    \caption{Pretrained MLLMs miss the small localized cues that define faceial DeepFakes. \methodname{} surfaces them by routing a dense-prediction stream into the language model alongside the contrastive stream.}
    \label{fig:missed_vs_captured}
\end{figure}

\noindent
\ul{\textbf{Localized Context:}} Contrastive training compresses per-patch detail in service of whole-image alignment, which is what produces the systematic blindness to small localized cues documented in Figure~\ref{fig:missed_vs_captured}: asymmetric eyes, unnaturally curved nose lines, faint blending boundaries. A model fluent on global semantics but blind on local patches will reliably miss the evidence that defines face-manipulation cues, even after task-specific finetuning (Table~\ref{tab:ablation_features}). We address this by routing a second stream into the language model from a VOM ($\Phi$) trained on dense patch prediction. Self-supervised dense prediction optimizes for patch-level distinctiveness rather than caption alignment, preserving the localized signal that contrastive training discards. $\Phi$ also operates by partitioning input $\mathcal{I}$ into non-overlapping $p\times p$ patches and produces,
\begin{equation}\label{eq:local}
    \mathcal{F}_{\Phi} = \Phi(\mathcal{I}^{H\times W\times 3}) \in \mathbb{R}^{n_t\times d_{\Phi}}.
\end{equation}
\noindent
These $n_t$ patch tokens are language-unaligned by construction ($\Phi$ has never been trained against text), which motivates the cross-modal adapter introduced in Sec.~\ref{subsec:mof}. Critically, both encoders operate on the same $p\times p$ patch grid: the $i$-th token in $\mathcal{F}_{\Omega}$ and the $i$-th token in $\mathcal{F}_{\Phi}$ describe the same spatial region. This shared grid is what makes spatial-correspondence-preserving fusion possible.

\subsection{Spatial-Correspondence-Preserving Fusion}\label{subsec:mof}
Fusing the two streams faces two structural challenges that distinguish this setting from generic multimodal fusion. First, the streams live on different manifolds by design: $\mathcal{F}_{\Omega}$ is aligned with the LLM's caption-trained embedding space, while $\mathcal{F}_{\Phi}$ is shaped by dense patch prediction and language-unaligned. Second, they describe the same image at the same spatial resolution: the $i$-th token of each stream corresponds to the same image patch (Sec.~\ref{subsec:foundation_models}). A fusion design that ignores either fact will perform suboptimally. We address each below.

\noindent
\ul{\textbf{Adapter: Cross-Manifold Projection.}} $\mathcal{F}_{\Omega}$ already lives in the LLM's input space and is consumed directly. $\mathcal{F}_{\Phi}$ requires a projection, but generic dimension matching is not sufficient: the adapter must move features from a manifold optimized for patch-level distinctiveness into a manifold optimized for whole-image alignment with captions, without collapsing the patch-level distinctions that motivated routing the VOM into the model in the first place. We instantiate $\mathcal{A}$ as a lightweight learnable map,
\begin{equation}\label{eq:adapter}
\mathcal{F}_{\mathcal{A}} = \mathcal{A}(\mathcal{F}_{\Phi}) \in \mathbb{R}^{n_t \times d_{\Omega}},
\end{equation}
\noindent
and observe that how $\mathcal{A}$ is trained matters more than how it is parameterized: the two-stage protocol in Sec.~\ref{sec:training_strategy} is what gives $\mathcal{A}$ its locality-preserving behavior, and without it the same architecture collapses (Table~\ref{tab:adapter_ablations}).

\noindent
\ul{\textbf{Token-Aligned Interleaving:}} With $\mathcal{F}_{\Omega}$ and $\mathcal{F}_{\mathcal{A}}$ in a common embedding space, the remaining question is how to combine them. The obvious choice is concatenation along the token dimension,
\begin{equation}\label{eq:concat}
\mathcal{F}_{\texttt{C}} = \mathcal{F}_{\Omega} \Vert \mathcal{F}_{\mathcal{A}} \in \mathbb{R}^{2n_t \times d_{\Omega}}.
\end{equation}
Concatenation is lossless, but it discards the structural fact from Sec.~\ref{subsec:foundation_models} that $\mathcal{F}_{\Omega}[i]$ and $\mathcal{F}_{\mathcal{A}}[i]$ describe the same spatial region. Under concatenation, this correspondence has to be recovered by the LLM through learned attention across a token distance of $n_t$ positions. A design that honors the shared grid by construction avoids this. We interleave the streams token by token,
\begin{align}
&\quad \mathcal{F}_{\texttt{I}}^k =
\begin{cases}
\mathcal{F}_{\Omega}^j, & k = 2j - 1 \\
\mathcal{F}_{\mathcal{A}}^j, & k = 2j
\end{cases} \\
&\quad\quad\quad\quad\small j \in \{1, 2, \ldots, n_t\}, k \in \{1, 2, \ldots, d_{\Omega}\} \notag
\end{align}
The contrastive and dense-prediction views of every spatial region now appear as adjacent tokens. The LLM no longer has to learn which global token corresponds to which local token; the answer is given by position. This design adds no learnable parameters, so the gain over concatenation in Table~\ref{tab:ablation_features} cannot be attributed to additional capacity: it is the inductive bias of the fusion operator itself.

\subsection{Disentangled Adapter Training}\label{sec:training_strategy}
The adapter $\mathcal{A}$ has two purposes: align $\mathcal{F}_{\Phi}$ with the LLM's input space (a general visual-linguistic problem), and preserve patch-level distinctiveness through the projection (a forensic problem, since it is this distinctiveness that surfaces face-manipulation cues). Optimizing both jointly on DeepFake data alone couples them: the adapter learns whichever projection minimizes loss on the training distribution, which is typically a shortcut that ties locality to a small set of manipulation patterns and does not generalize. We decouple the two goals by training in two stages. Stage~1 establishes general visual-linguistic alignment before any DeepFake data is seen, and Stage~2 specializes the now-aligned adapter and LLM jointly on the forensic task. Table~\ref{tab:adapter_ablations} confirms that removing either stage, or collapsing them into a single joint run, degrades performance significantly.

\noindent
\ul{\textbf{Visual Representation Alignment:}} We train $\mathcal{A}$ on the LCS-558K image-caption corpus~\cite{liu2024visual} with the LLM frozen, optimizing $\mathcal{L}_{\text{LLM}}$ (defined below) over standard caption generation. This stage sees no DeepFake data, no forensic vocabulary, and no region-level supervision. Its only objective is to teach $\mathcal{A}$ how to map $\mathcal{F}_{\Phi}$ into a space the language model can consume, using whatever visual concepts a generic caption corpus contains. The result is an adapter aligned to language at the manifold level, before it is asked to surface any task-specific cue. Skipping this stage (\emph{No general pre-adaptation} in Table~\ref{tab:adapter_ablations}) collapses cross-domain performance, because $\mathcal{A}$ has no general visual-linguistic alignment to fall back on when the task distribution shifts.

\noindent
\ul{\textbf{Task-Specific Grounding:}} With $\mathcal{A}$ now language-aligned, we unfreeze the LLM and jointly optimize $\mathcal{A}$ and the LLM on language-annotated DeepFake data. $\mathcal{A}$ remains trainable rather than frozen: freezing it (\emph{No task-adaptive refinement} in Table~\ref{tab:adapter_ablations}) leaves the adapter unable to specialize its projection for forensic attributes, and forensic precision drops accordingly. The joint update lets the adapter sharpen its locality-preservation in directions that matter for the task while the LLM learns the corresponding linguistic mapping. Collapsing Stage~1 and Stage~2 into a single joint run on DeepFake data (\emph{Joint training only}) achieves a middle outcome: the adapter has no foundational alignment to anchor it, so its task specialization is tied to manipulation patterns in the training set and transfers poorly across content categories.

\noindent
\ul{\textbf{Loss Objective:}} Both stages share the same autoregressive language-modeling loss. For each example, the question $\mathcal{Q}$ and reference answer $\mathcal{E}$ are tokenized as
\begin{align}
\mathcal{T}_{\text{prefix}} = \tau(\mathcal{Q})&, \quad \mathcal{T}_{\text{separator}} = \tau(\text{``}\backslash n\text{''}), \\\quad \mathcal{T}_{\text{suffix}} = \tau(\mathcal{E})&, \notag
\mathcal{T} = \mathcal{T}_{\text{prefix}} \Vert \mathcal{T}_{\text{separator}} \Vert \mathcal{T}_{\text{suffix}}.
\end{align}
Cross-entropy is computed only over answer tokens, enforced by a loss mask $\mathcal{M}_{\text{loss}} = [0]^{|\mathcal{T}_{\text{prefix}}| + |\mathcal{T}_{\text{separator}}|} \Vert [1]^{|\mathcal{T}_{\text{suffix}}|}$. $\mathcal{T}_{\text{prefix}}$ attend bidirectionally and $\mathcal{T}_{\text{suffix}}$ attend causally to support autoregressive generation. The objective is, thus,
\begin{equation}
\mathcal{L}_{\text{LLM}} = -\frac{\sum_{i=1}^{|\mathcal{T}|} \mathcal{M}_{\text{loss}}^i \cdot \log P(\mathcal{T}^i)}{\sum_{i=1}^{|\mathcal{T}|} \mathcal{M}_{\text{loss}}^i},
\end{equation}
where $P(\cdot)$ is the model's predicted distribution over tokens. At inference, only $\mathcal{T}_{\text{prefix}}$ is provided and the model generates responses autoregressively.
\section{Experiments}\label{sec:results}

\noindent
\ul{\textbf{Training \& Evaluation Protocol:}} \methodname{} is trained on DD-VQA \cite{zhang2024common} (face-manipulated FF++ \cite{rossler2019faceforensics++} with language annotations) and DVF \cite{song2024learning} (fully synthetic, language-annotated). Detection performance is also evaluated cross-domain on CelebDF \cite{li2020celeb} and DF40 \cite{yan2024df40} (face-manipulated) and DeMamba \cite{chen2024demamba} (fully synthetic), which lack language annotations. Dataset details are in \S\ref{suppl:dataset_details}; implementation details (hardware/software) in \S\ref{suppl:implementation}.

For each example, we report (i) an \emph{LLM-as-a-judge accuracy} that uses Gemini-1.0 \cite{team2023gemini} to check whether the model's verdict (real vs.~fake) agrees with the reference; and (ii) standard generation metrics BLEU-3, BLEU-4, ROUGE-L, and CIDEr against the reference explanation, following \cite{chen2024x2dfd, zhang2024common}. These two families of metrics are necessary but not sufficient. The binary judge captures whether conclusions agree, not whether the supporting reasoning is faithful, and $n$-gram metrics correlate weakly with human judgments of explanation quality \cite{novikova2017we}. We therefore complement them with a dedicated explanation-quality evaluation: a three-annotator human study on 300 stratified samples scoring faithfulness, specificity, and forensic helpfulness; a rubric-based GPT-4o judge calibrated against the human ratings (Pearson $r \geq 0.76$ on every dimension) and applied at full scale; and a faithfulness analysis that quantifies hallucinated region and attribute claims. Full protocol, instructions, and detailed results are in \S\ref{sec:explanation_eval}.

\begin{table}[]
\centering
\caption{\textbf{SOTA Comparison}. The competitor methods are specifically designed for either face-manipulation or synthetic data reasoning. However our \methodname{} framework is designed for generalized reasoning on both face-manipulated and synthetic content, and still outperforms in both categories. \colorbox{\red{}}{\textbf{Best}} and \colorbox{\blue{}}{second-best} performances are marked. $^*$-marked methods were run using official implementation and model weights.}
\vspace{-1em}
\resizebox{\columnwidth}{!}{
\begin{tabular}{c|c|c|c|c|c|c}
\hline
\textbf{Dataset}          & \textbf{Method}                                             & \textbf{Accuracy} & \textbf{AUC}     & \textbf{BLEU-4} & \textbf{ROUGE-L} & \textbf{CIDEr}  \\ \hline
\multicolumn{7}{c}{\textit{Face Manipulated Data}}\\ \hline
\multirow{14}{*}{\begin{tabular}[c]{@{}c@{}}DD-VQA\\ \cite{zhang2024common} \end{tabular}}  & XceptionNet-BLIP-TI \cite{zhang2024common} & 89.25\%           & 92.24\%                & -               & -                 & -               \\  
                          & HifiNet \cite{guo2023hierarchical}         & 89.16\%           & 92.10\%                & -               & -                 & -               \\  
                          & HifiNet-BLIP-TI \cite{zhang2024common}     & 91.25\%           & 95.14\%                & -               & -                 & -               \\  
                          & RECCE \cite{cao2022end}                    & 91.03\%           & 95.02\%                & -               & -                 & -               \\  
                          & RECCE-BLIP \cite{zhang2024common}          & 89.22\%           & 93.71\%                & -               & -                 & -               \\  
                          & RECCE-BLIP-TI \cite{zhang2024common}       & \cellcolor{\blue{}}{92.08\%}     & \cellcolor{\blue{}}{95.36\%}                & -               & -                 & -               \\  
                          & BLIP \cite{li2022blip}                     & 81.68\%           & -                & 0.3569          & 0.5664            & 1.8177          \\  
                          & BLIP-T \cite{zhang2024common}              & 83.65\%           & -                & 0.3714          & 0.5774            & 1.8715          \\  
                          & BLIP-I \cite{zhang2024common}              & 84.87\%           & -                & 0.3800          & 0.5882            & 1.8931          \\  
                          & BLIP-TI \cite{zhang2024common}             & 87.49\%           & -                & \cellcolor{\blue{}}{0.4075}    & \cellcolor{\blue{}}{0.6085}      & \cellcolor{\blue{}}{2.0567}    \\
                          & $\mathcal{X}^2$-DFD \cite{chen2024x2dfd}   & -                 & -                & 0.2030          & 0.1550            & 0.0270          \\
                          & ForgerySleuth$^*$ \cite{sun2025forgerysleuth}  & 89.42\%           & 93.18\%          & 0.3812          & 0.5854            & 1.9486          \\  
                          & FakeShield$^*$ \cite{xu2025fakeshield}         & 90.85\%           & 93.91\%          & 0.3947          & 0.5982            & 2.0143          \\  \arrayrulecolor[gray]{0.85} \cline{2-7} \arrayrulecolor{black}
                          & \textbf{\methodname{}}                                               & \cellcolor{\red{}}{\textbf{94.12\%}}  & \cellcolor{\red{}}{\textbf{95.39\%}} & \cellcolor{\red{}}{\textbf{0.4304}} & \cellcolor{\red{}}{\textbf{0.6285} }  & \cellcolor{\red{}}{\textbf{2.6321}} \\ \hline
\multicolumn{7}{c}{\textit{Synthetic Data}}\\ \hline
\multirow{13}{*}{\begin{tabular}[c]{@{}c@{}}DVF\\ \cite{song2024learning}\end{tabular}}    & CNNDet \cite{wang2020cnn}               & -                 & 78.20\%          & -                & -                 & -               \\  
                                                                                                            & DIRE \cite{wang2023dire}                & -                 & 62.10\%          & -                & -                 & -               \\  
                                                                                                            & \cite{cozzolino2024raising}     & -                 & 67.00\%          & -                & -                 & -               \\  
                                                                                                            & UNI-FD \cite{ojha2023towards}           & -                 & 74.10\%          & -                & -                 & -               \\  
                                                                                                            & F3Net \cite{qian2020thinking}           & -                 & 81.30\%          & -                & -                 & -               \\  
                                                                                                            & ViViT \cite{arnab2021vivit}             & -                 & 79.10\%          & -                & -                 & -               \\  
                                                                                                            & TALL \cite{xu2023tall}                  & -                 & 69.50\%          & -                & -                 & -               \\  
                                                                                                            & TS2-Net \cite{liu2022ts2}               & -                 & 72.10\%          & -                & -                 & -               \\  
                                                                                                            & DE-FAKE \cite{sha2023fake}              & -                 & 72.10\%          & -                & -                 & -               \\  
                                                                                                            & HifiNet \cite{guo2023hierarchical}      & -                 & 84.30\%          & -                & -                 & -               \\  
                                                                                                            & DVF \cite{song2024learning}             & -                 & \cellcolor{\blue{}}{92.00\%}    & -                & -                 & -               \\ 
                                                                                                            & FakeShield$^*$ \cite{xu2025fakeshield}      & 88.74\%           & 89.83\%          & \cellcolor{\blue{}}{0.3924}    & \cellcolor{\blue{}}{0.4985}      & \cellcolor{\blue{}}{1.9482}    \\ \arrayrulecolor[gray]{0.85} \cline{2-7} \arrayrulecolor{black} 
                                                                                                            & \textbf{\methodname{}}                  & \cellcolor{\red{}}{\textbf{94.47\%}}  & \cellcolor{\red{}}{\textbf{95.82\%}} & \cellcolor{\red{}}{\textbf{0.4279}}  & \cellcolor{\red{}}{\textbf{0.5338}}   & \cellcolor{\red{}}{\textbf{2.1744}} \\ \hline
\end{tabular}
}
\label{tab:sota_comparison_nlg}
\vspace{-1em}
\end{table}

\noindent
\ul{\textbf{State-of-the-art Comparison:}} Table~\ref{tab:sota_comparison_nlg} compares \methodname{} against baselines on DD-VQA (face-manipulated) and DVF (fully synthetic). MLLM-based reasoning baselines exist only on the face-manipulated side (BLIP-TI, X${}^2$-DFD, ForgerySleuth), reflecting the literature bifurcation; on the synthetic side, our reasoning-metric competitors are detection-only methods. The single baseline that operates on both regimes is FakeShield~\cite{xu2025fakeshield}, which scores reasonably on DD-VQA (90.85\% accuracy, 2.01 CIDEr) but lags on DVF (88.74\% accuracy, 89.83\% AUC, 1.95 CIDEr), illustrating that near-unified treatments still leave a coverage gap. \methodname{} leads on every metric on both datasets: +3.27\% accuracy and +0.62 CIDEr over FakeShield on DD-VQA, and +5.73\% accuracy, +5.99\% AUC, and +0.23 CIDEr on DVF. Cross-domain detection results on CelebDF, DF40, and DeMamba (in \S\ref{suppl:detection_performance}) show \methodname{} outperforming competitors evaluated in-domain despite being itself evaluated purely cross-domain.

\begin{figure*}
    \centering
    \includegraphics[width=\textwidth]{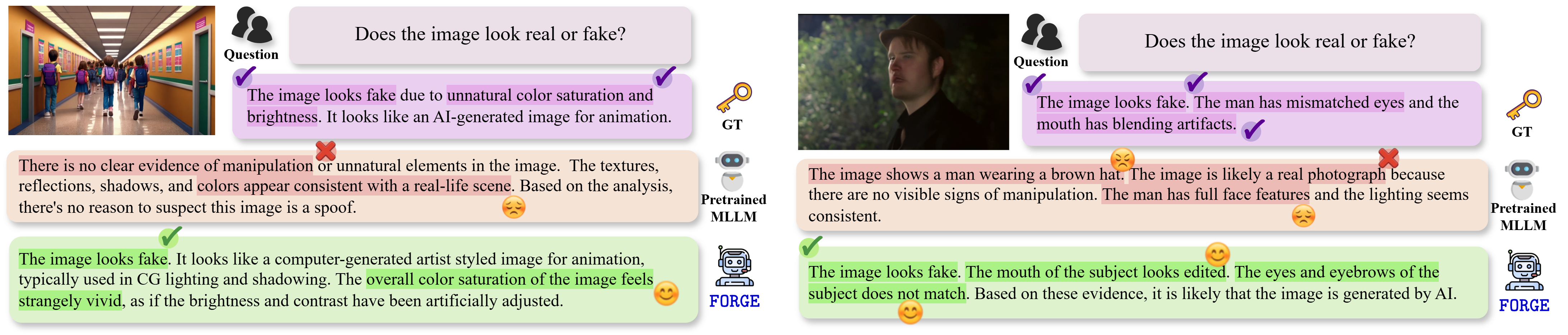}
    \caption{Examples demonstrating the DeepFake relevance of the responses generated by \methodname{} compared to a pretrained MLLM (PaliGemma2). \methodname{} not only enhances accuracy on DeepFake detection, but also improves relevance to the task providing finegrained reasoning. Additional examples in \S\ref{suppl:qualitative}}
    \label{fig:qualitative_results}
\end{figure*}

\noindent
\ul{\textbf{Architectural Ablations:}} Table~\ref{tab:ablation_features} walks five configurations from the off-the-shelf MLLM up to the full \methodname{} system, each falsifying a successively stronger ``simpler explanation'' for the gains. The off-the-shelf MLLM scores 11.50\% accuracy on DD-VQA and 33.07\% on DVF, well below the 50\% binary baseline and consistent with the zero-shot failures reported in \cite{jia2024can}: scale alone does not solve the task. Finetuning the same MLLM on DeepFake reasoning data with no second stream (\emph{naive grounding}, the $\mathcal{F}_{\Omega}$-only row) lifts performance to 88.19\% / 79.88\%; adding the VOM stream is what closes the remaining gap to the full system, contributing +5.9 / +14.6 accuracy points on DD-VQA / DVF. The single-stream VOM baseline ($\mathcal{F}_{\mathcal{A}}$-only) shows the opposite asymmetry: stronger on DD-VQA (90.45\%, where face manipulations are localized patches) and weaker on DVF (77.67\%, where scene-level cues matter), confirming that neither inductive bias is sufficient alone. Combining the two streams via concatenation ($\mathcal{F}_{\texttt{C}}$) closes most of the gap to 93.56\% / 92.81\%, but token-aligned interleaving ($\mathcal{F}_{\texttt{I}}$, ours) extracts a further 0.6-1.7 points on the same data, the same encoders, and the same training pipeline. Since interleaving adds zero learnable parameters, this final gain cannot be attributed to capacity; it is the spatial alignment of the fusion operator itself, as argued in Sec. \ref{subsec:mof}.

\begin{table}[]
\centering
\caption{\textbf{Architectural ablations on \methodname{}'s vision pathway.} From a pretrained MLLM (capacity baseline), through single-stream and dual-stream variants, to spatial-correspondence-preserving fusion. \colorbox{\red{}}{\textbf{Best}} and \colorbox{\blue{}}{second-best} performances are marked.}
\vspace{-1em}
\resizebox{\columnwidth}{!}{
\begin{tabular}{c|c|c|c|c|c|c}
\hline
\textbf{Dataset}        & \textbf{Configuration}    & \textbf{Accuracy} $\uparrow$ & \textbf{BLEU-3} $\uparrow$ & \textbf{BLEU-4} $\uparrow$ & \textbf{ROUGE-L} $\uparrow$ & \textbf{CIDEr} $\uparrow$  \\ \hline
\multirow{5}{*}{\begin{tabular}[c]{@{}c@{}}DD-VQA\\ \cite{zhang2024common}\end{tabular}} 
                        & Pretrained MLLM                                          & 11.50\%           & 0.0000           & 0.0000           & 0.0218            & 0.2610          \\
                        & MLLM only ($\mathcal{F}_{\Omega}$)                       & 88.19\%           & 0.4012           & 0.3768           & 0.5633            & 2.1478          \\
                        & VOM only ($\mathcal{F}_{\mathcal{A}}$)                   & 90.45\%           & 0.4107           & 0.4143           & 0.5843            & 2.0816          \\
                        & \methodname{} w/ concat. ($\mathcal{F}_{\texttt{C}}$)    & \cellcolor{\blue{}}{93.56\%}           & \cellcolor{\blue{}}{0.4497}           & \cellcolor{\blue{}}{0.4286}           & \cellcolor{\blue{}}{0.6015}            & \cellcolor{\blue{}}{2.4193}          \\ 
                        & \methodname{} ($\mathcal{F}_{\texttt{I}}$)              & \cellcolor{\red{}}{\textbf{94.12\%}}  & \cellcolor{\red{}}{\textbf{0.4649}}  & \cellcolor{\red{}}{\textbf{0.4304}}  & \cellcolor{\red{}}{\textbf{0.6285}}   & \cellcolor{\red{}}{\textbf{2.6321}} \\ \hline
\multirow{5}{*}{\begin{tabular}[c]{@{}c@{}}DVF\\ \cite{song2024learning}\end{tabular}}    
                        & Pretrained MLLM                                          & 33.07\%           & 0.0000           & 0.0000           & 0.0034            & 0.0001          \\
                        & MLLM only ($\mathcal{F}_{\Omega}$)                       & 79.88\%           & 0.3142           & 0.3839           & 0.4877            & 1.8629          \\ 
                        & VOM only ($\mathcal{F}_{\mathcal{A}}$)                   & 77.67\%           & 0.3077           & 0.3786           & 0.4873            & 1.7465          \\ 
                        & \methodname{} w/ concat. ($\mathcal{F}_{\texttt{C}}$)    & \cellcolor{\blue{}}{92.81\%}           & \cellcolor{\blue{}}{0.3614}           & \cellcolor{\blue{}}{0.4261}           & \cellcolor{\blue{}}{0.5234}            & \cellcolor{\blue{}}{2.1605}          \\ 
                        & \methodname{} ($\mathcal{F}_{\texttt{I}}$)              & \cellcolor{\red{}}{\textbf{94.47\%}}  & \cellcolor{\red{}}{\textbf{0.3786}}  & \cellcolor{\red{}}{\textbf{0.4279}}  & \cellcolor{\red{}}{\textbf{0.5338}}   & \cellcolor{\red{}}{\textbf{2.1744}} \\ \hline
\end{tabular}
}
\label{tab:ablation_features}
\end{table}

\begin{table}[]
\centering
\caption{\textbf{Ablation on Adapter Configurations.}}
\vspace{-1em}
\resizebox{\columnwidth}{!}{
\begin{tabular}{ccccccc}
\hline
\multicolumn{1}{c|}{\textbf{Dataset}}         & \multicolumn{1}{c|}{\textbf{Adapter Settings}}                                                      & \multicolumn{1}{c|}{\textbf{Accuracy} $\uparrow$} & \multicolumn{1}{c|}{\textbf{BLEU-3} $\uparrow$} & \multicolumn{1}{c|}{\textbf{BLEU-4} $\uparrow$} & \multicolumn{1}{c|}{\textbf{ROUGE-L} $\uparrow$} & \textbf{CIDEr} $\uparrow$                                                                                                        \\ \hline
\multicolumn{7}{c}{\textit{Face-Manipulated Data}}                                                                                                                                                                                                                                                                                                                                                                                                                                                                                                                                                                                                                                                                                                                                                                                                               \\ \hline
\multicolumn{1}{c|}{\multirow{4}{*}{\begin{tabular}[c]{@{}c@{}}DD-VQA\\ \cite{zhang2024common}\end{tabular}}}  & \multicolumn{1}{c|}{No general pre-adaptation}     & \multicolumn{1}{c|}{79.13\%}                                                                                                                 & \multicolumn{1}{c|}{0.3017}                                                                                                                & \multicolumn{1}{c|}{0.2975}                                                                                                                & \multicolumn{1}{c|}{0.5113}                                                                                                                & 1.5812                                                                                                                \\ \arrayrulecolor[gray]{0.85} \cline{2-7} \arrayrulecolor{black} 
\multicolumn{1}{c|}{}                         & \multicolumn{1}{c|}{No task-adaptive refinement}                                                                 & \multicolumn{1}{c|}{\begin{tabular}[c]{@{}c@{}}86.56\%\\ {\color[HTML]{009901}{(\(\uparrow\)7.43\%)}}\end{tabular}}  & \multicolumn{1}{c|}{\begin{tabular}[c]{@{}c@{}}0.3964\\ {\color[HTML]{009901}{(\(\uparrow\)0.0947)}}\end{tabular}} & \multicolumn{1}{c|}{\begin{tabular}[c]{@{}c@{}}0.3897\\ {\color[HTML]{009901}{(\(\uparrow\)0.0922)}}\end{tabular}} & \multicolumn{1}{c|}{\begin{tabular}[c]{@{}c@{}}0.5712\\ {\color[HTML]{009901}{(\(\uparrow\)0.0599)}}\end{tabular}} & \begin{tabular}[c]{@{}c@{}}2.0447\\ {\color[HTML]{009901}{(\(\uparrow\)0.4635)}}\end{tabular} \\ \arrayrulecolor[gray]{0.85} \cline{2-7} \arrayrulecolor{black} 
\multicolumn{1}{c|}{}                         & \multicolumn{1}{c|}{Joint training only} & \multicolumn{1}{c|}{\begin{tabular}[c]{@{}c@{}}90.371\%\\ {\color[HTML]{009901}{(\(\uparrow\)3.81\%)}}\end{tabular}} & \multicolumn{1}{c|}{\begin{tabular}[c]{@{}c@{}}0.4107\\ {\color[HTML]{009901}{(\(\uparrow\)0.0143)}}\end{tabular}} & \multicolumn{1}{c|}{\begin{tabular}[c]{@{}c@{}}0.4066\\ {\color[HTML]{009901}{(\(\uparrow\)0.0169)}}\end{tabular}} & \multicolumn{1}{c|}{\begin{tabular}[c]{@{}c@{}}0.6015\\ {\color[HTML]{009901}{(\(\uparrow\)0.0303)}}\end{tabular}} & \begin{tabular}[c]{@{}c@{}}2.3186\\ {\color[HTML]{009901}{(\(\uparrow\)0.2739)}}\end{tabular} \\ \arrayrulecolor[gray]{0.85} \cline{2-7} \arrayrulecolor{black} 
\multicolumn{1}{c|}{}                         & \multicolumn{1}{c|}{\textbf{\methodname{}}}                                                                  & \multicolumn{1}{c|}{\begin{tabular}[c]{@{}c@{}}\textbf{94.12\%}\\ {\color[HTML]{009901}{(\(\uparrow\)3.75\%)}}\end{tabular}}  & \multicolumn{1}{c|}{\begin{tabular}[c]{@{}c@{}}\textbf{0.4649}\\ {\color[HTML]{009901}{(\(\uparrow\)0.0542)}}\end{tabular}} & \multicolumn{1}{c|}{\begin{tabular}[c]{@{}c@{}}\textbf{0.4304}\\ {\color[HTML]{009901}{(\(\uparrow\)0.0238)}}\end{tabular}} & \multicolumn{1}{c|}{\begin{tabular}[c]{@{}c@{}}\textbf{0.6285}\\ {\color[HTML]{009901}{(\(\uparrow\)0.0270)}}\end{tabular}} & \begin{tabular}[c]{@{}c@{}}\textbf{2.6321}\\ {\color[HTML]{009901}{(\(\uparrow\)0.3135)}}\end{tabular} \\ \hline
\multicolumn{7}{c}{\textit{Synthetic Data}}                                                                                                                                                                                                                                                                                                                                                                                                                                                                                                                                                                                                                                                                                                                                                                                                                      \\ \hline
\multicolumn{1}{c|}{\multirow{4}{*}{\begin{tabular}[c]{@{}c@{}}DVF\\ \cite{song2024learning}\end{tabular}}}     & \multicolumn{1}{c|}{No general pre-adaptation}     & \multicolumn{1}{c|}{74.39\%}                                                                                                                 & \multicolumn{1}{c|}{0.2317}                                                                                                                & \multicolumn{1}{c|}{0.2815}                                                                                                                & \multicolumn{1}{c|}{0.4014}                                                                                                                & 1.0961                                                                                                                \\ \arrayrulecolor[gray]{0.85} \cline{2-7} \arrayrulecolor{black} 
\multicolumn{1}{c|}{}                         & \multicolumn{1}{c|}{No task-adaptive refinement}                                                                 & \multicolumn{1}{c|}{\begin{tabular}[c]{@{}c@{}}89.23\%\\ {\color[HTML]{009901}{(\(\uparrow\)14.84\%)}}\end{tabular}} & \multicolumn{1}{c|}{\begin{tabular}[c]{@{}c@{}}0.3173\\ {\color[HTML]{009901}{(\(\uparrow\)0.0856)}}\end{tabular}} & \multicolumn{1}{c|}{\begin{tabular}[c]{@{}c@{}}0.3681\\ {\color[HTML]{009901}{(\(\uparrow\)0.0866)}}\end{tabular}} & \multicolumn{1}{c|}{\begin{tabular}[c]{@{}c@{}}0.4846\\ {\color[HTML]{009901}{(\(\uparrow\)0.0832)}}\end{tabular}} & \begin{tabular}[c]{@{}c@{}}1.6820\\ {\color[HTML]{009901}{(\(\uparrow\)0.5859)}}\end{tabular} \\ \arrayrulecolor[gray]{0.85} \cline{2-7} \arrayrulecolor{black} 
\multicolumn{1}{c|}{}                         & \multicolumn{1}{c|}{Joint training only} & \multicolumn{1}{c|}{\begin{tabular}[c]{@{}c@{}}92.11\%\\ {\color[HTML]{009901}{(\(\uparrow\)2.88\%)}}\end{tabular}}  & \multicolumn{1}{c|}{\begin{tabular}[c]{@{}c@{}}0.3350\\ {\color[HTML]{009901}{(\(\uparrow\)0.0177)}}\end{tabular}}  & \multicolumn{1}{c|}{\begin{tabular}[c]{@{}c@{}}0.3875\\ {\color[HTML]{009901}{(\(\uparrow\)0.0194)}}\end{tabular}} & \multicolumn{1}{c|}{\begin{tabular}[c]{@{}c@{}}0.4972\\ {\color[HTML]{009901}{(\(\uparrow\)0.0126)}}\end{tabular}} & \begin{tabular}[c]{@{}c@{}}1.8988\\ {\color[HTML]{009901}{(\(\uparrow\)0.2168)}}\end{tabular} \\ \arrayrulecolor[gray]{0.85} \cline{2-7} \arrayrulecolor{black} 
\multicolumn{1}{c|}{}                         & \multicolumn{1}{c|}{\textbf{\methodname{}}}                                                                  & \multicolumn{1}{c|}{\begin{tabular}[c]{@{}c@{}}\textbf{94.47\%}\\ {\color[HTML]{009901}{(\(\uparrow\)2.36\%)}}\end{tabular}}  & \multicolumn{1}{c|}{\begin{tabular}[c]{@{}c@{}}\textbf{0.3786}\\ {\color[HTML]{009901}{(\(\uparrow\)0.0436)}}\end{tabular}} & \multicolumn{1}{c|}{\begin{tabular}[c]{@{}c@{}}\textbf{0.4279}\\ {\color[HTML]{009901}{(\(\uparrow\)0.0404)}}\end{tabular}} & \multicolumn{1}{c|}{\begin{tabular}[c]{@{}c@{}}\textbf{0.5338}\\ {\color[HTML]{009901}{(\(\uparrow\)0.0366)}}\end{tabular}} & \begin{tabular}[c]{@{}c@{}}\textbf{2.1744}\\ {\color[HTML]{009901}{(\(\uparrow\)0.2756)}}\end{tabular} \\ \hline
\end{tabular}
}
\label{tab:adapter_ablations}
\end{table}

\begin{table}[t]
\centering
\caption{\textbf{Faithfulness and hallucination analysis.} RHR: Region Hallucination Rate (fraction of explanations referencing an authentic region as manipulated). AHR: Attribute Hallucination Rate (fraction of attribute claims unsupported by ground-truth attribute labels). FP: Forensic Precision (fraction of attribute claims on fake images that align with annotated manipulation cues). \methodname{} cuts RHR by more than $2\times$ over the strongest baseline on both datasets.}
\vspace{-1em}
\resizebox{\linewidth}{!}{
\begin{tabular}{llccc}
\hline
Dataset & Method & RHR $\downarrow$ & AHR $\downarrow$ & FP $\uparrow$ \\
\hline
\multirow{6}{*}{DD-VQA~\cite{zhang2024common}}
 & Pretrained PaliGemma2~\cite{steiner2024paligemma} & 64.3\% & 71.2\% & 18.4\% \\
 & ForgerySleuth~\cite{sun2025forgerysleuth}         & 39.7\% & 43.8\% & 50.1\% \\
 & FakeShield~\cite{xu2025fakeshield}                & 36.4\% & 40.2\% & 53.6\% \\
 & BLIP-TI~\cite{zhang2024common}                    & 41.2\% & 45.6\% & 48.7\% \\
 & Grounded PaliGemma2 (no VOM)                       & 28.9\% & 31.5\% & 67.8\% \\
 & \textbf{\methodname{}}                          & \textbf{12.7\%} & \textbf{14.3\%} & \textbf{84.6\%} \\
\hline
\multirow{5}{*}{DVF~\cite{song2024learning}}
 & Pretrained PaliGemma2~\cite{steiner2024paligemma} & 58.7\% & 66.4\% & 22.1\% \\
 & FakeShield~\cite{xu2025fakeshield}                & 38.6\% & 42.1\% & 51.8\% \\
 & MM-Det~\cite{song2024learning}                    & 35.2\% & 38.9\% & 55.4\% \\
 & Grounded PaliGemma2 (no VOM)                       & 30.1\% & 33.8\% & 64.2\% \\
 & \textbf{\methodname{}}                          & \textbf{15.4\%} & \textbf{17.2\%} & \textbf{81.3\%} \\
\hline
\end{tabular}
}
\label{tab:faithfulness}
\vspace{-1em}
\end{table}

\noindent
\ul{\textbf{Adapter Ablation:}} Table~\ref{tab:adapter_ablations} tests the two-stage training protocol against three alternatives: skipping the foundational alignment stage (adapter trained only on DeepFake data; \emph{No general pre-adaptation}), skipping the task-specific stage (adapter pretrained on LCS-558K but frozen during downstream training; \emph{No task-adaptive refinement}), and collapsing the two stages into one (adapter and LLM jointly optimized on DeepFake data from scratch; \emph{Joint training only}). Each maps to a specific prediction from Sec. \ref{sec:training_strategy}, and each underperforms in the way that prediction implies. Removing the foundational stage costs the most: 79.13\% / 74.39\% on DD-VQA / DVF, the largest gap to the full system, because the adapter has no general visual-linguistic alignment to fall back on when the task distribution shifts. Freezing the adapter after the foundational stage reaches 86.56\% / 89.23\%; the alignment is in place but the adapter cannot sharpen its projection for forensic attributes. Collapsing the two stages reaches 90.37\% / 92.11\%, the closest to \methodname{} and the most informative comparison: the adapter has access to all the same data but couples the alignment and specialization jobs, and the remaining 2-4\% gap is what disentangling them recovers.

\begin{table}[]
\centering
\caption{\textbf{VOM ablation:} Replacing DINOv2 with MAE (ViT-L/14) isolates the contribution of dense patch prediction as a training paradigm from the specific choice of encoder. Both dense-prediction encoders substantially exceed the contrastive-only baseline (row-1), and DINOv2's small residual advantage is consistent with feature-space over pixel-space reconstruction.}
\vspace{-1em}
\resizebox{\columnwidth}{!}{
\begin{tabular}{c|c|c|c|c|c|c|c}
\hline
\multirow{2}{*}{\textbf{VOM ($\Phi$)}} & \multicolumn{3}{c|}{\textbf{DD-VQA} \cite{zhang2024common}} & \multicolumn{3}{c|}{\textbf{DVF} \cite{song2024learning}} & \multirow{2}{*}{\textbf{Params}} \\ \cline{2-7}
                                        & \textbf{BLEU-4} $\uparrow$ & \textbf{ROUGE-L} $\uparrow$ & \textbf{CIDEr} $\uparrow$ & \textbf{BLEU-4} $\uparrow$ & \textbf{ROUGE-L} $\uparrow$ & \textbf{CIDEr} $\uparrow$ & \\ \hline
$\mathcal{F}_{\Omega}$ only (no VOM)    & 0.3768           & 0.5633           & 2.1478          & 0.3839           & 0.4877           & 1.8629          & --      \\
MAE ViT-L/14                            & 0.4143           & 0.6104           & 2.5391          & 0.4142           & 0.5187           & 2.1105          & 304M    \\
DINOv2 ViT-L/14 (ours)                  & \textbf{0.4304}  & \textbf{0.6285}  & \textbf{2.6321} & \textbf{0.4279}  & \textbf{0.5338}  & \textbf{2.1744} & 304M    \\ \hline
\end{tabular}
}
\label{tab:vom_ablation}
\vspace{-1em}
\end{table}

\noindent
\ul{\textbf{VOM Ablation:}} The architectural argument in Sec.~\ref{subsec:foundation_models} attributes the second stream's contribution to dense patch prediction as a training paradigm, not to DINOv2 specifically. Table~\ref{tab:vom_ablation} tests this claim by replacing DINOv2 with MAE (ViT-L/14, ImageNet-pretrained), keeping every other component of \methodname{} identical. MAE shares DINOv2's patch-level supervision paradigm (masked patch reconstruction) but reconstructs in pixel space rather than feature space and lacks DINOv2's multi-crop student-teacher stabilization. MAE recovers most of DINOv2's gain over the contrastive-only $\mathcal{F}_{\Omega}$ baseline (within 2-4\% relative on every metric on both datasets), and both dense-prediction encoders substantially exceed $\mathcal{F}_{\Omega}$ alone. This confirms that the gain comes from the dense patch prediction paradigm itself: any encoder whose training objective preserves per-patch distinctiveness supplies the localized signal the contrastive encoder discards. DINOv2's small residual advantage over MAE is consistent with feature-space reconstruction preserving semantic patch distinctions that pixel-space reconstruction does not, but the effect is second-order compared to the paradigm switch from contrastive to dense-prediction training.

\begin{table}[]
\centering
\caption{\textbf{Fusion strategy ablation:} Alternative parameterized fusion mechanisms compared to token-aligned interleaving on DD-VQA, holding all other components of \methodname{} fixed. Interleaving matches or outperforms every parameterized alternative while adding zero learnable parameters.}
\vspace{-1em}
\resizebox{\columnwidth}{!}{
\begin{tabular}{l|c|c|c|c}
\hline
\textbf{Fusion Strategy}    & \textbf{Added Params} & \textbf{Accuracy} $\uparrow$ & \textbf{BLEU-4} $\uparrow$ & \textbf{CIDEr} $\uparrow$ \\ \hline
Q-Former                    & $\sim$12M      & 93.10\%          & 0.4190           & 2.5100          \\
Cross-attention             & $\sim$4M       & 93.80\%          & 0.4270           & 2.5800          \\
Gated fusion                & $\sim$1M       & 93.70\%          & 0.4250           & 2.5900          \\
Interleaving (ours)         & \textbf{0}     & \textbf{94.12\%} & \textbf{0.4304}  & \textbf{2.6321} \\ \hline
\end{tabular}
}
\label{tab:fusion_ablation}
\vspace{-2em}
\end{table}

\noindent
\ul{\textbf{Fusion Strategy Ablation:}} Table~\ref{tab:fusion_ablation} compares token-aligned interleaving against three parameterized alternatives on DD-VQA: cross-attention (learned key/value projections between streams), Q-Former (compressive cross-modal projection), and gated fusion (per-token learned gates). All three underperform interleaving despite adding capacity, and Q-Former, the fusion strategy behind BLIP-2 and InstructBLIP, underperforms most. This isolates the contribution to the fusion operator's inductive bias rather than to any learned combination: preserving spatial correspondence by construction beats learning to recover it.

\noindent
\ul{\textbf{Region-Referential Reasoning:}} Detection accuracy can be high for the wrong reasons, so we directly evaluate whether \methodname{}'s explanations are faithful to the image and specific enough to be forensically useful. Three annotators (calibrated with a 90-minute domain expert session) rate 300 stratified samples on faithfulness, specificity, and forensic helpfulness (1-5 Likert; Krippendorff's $\alpha \geq 0.68$; full protocol in \S\ref{sec:explanation_eval}). Table~\ref{tab:human_eval} reports the results. \methodname{} leads every baseline on every dimension on both datasets, with the largest gains on specificity and forensic helpfulness (+0.77 and +0.65 over the grounded PaliGemma2 ablation on DD-VQA, with comparable margins on DVF), the dimensions most directly tied to region-referential reasoning. We complement perceived quality with an objective check: Table~\ref{tab:faithfulness} reports the rate at which generated explanations hallucinate manipulated regions (RHR) or unsupported attribute claims (AHR), and the precision of attribute claims on manipulated images (FP). \methodname{} reaches RHR of 12.7\% / 15.4\% on DD-VQA / DVF, less than half the next-best baseline (28.9\% / 30.1\%), and lifts forensic precision past 80\% on both datasets. The grounded PaliGemma2 ablation, which removes only the VOM stream, retains the higher hallucination rates, confirming that detection-finetuning alone does not produce faithful reasoning. A per-region breakdown across face attributes (eyes, nose, mouth, etc.) and scene aspects (lighting, shadows, textures, etc.) is in \S\ref{suppl:stratified}.

\noindent
\ul{\textbf{Qualitative Results:}} Figure~\ref{fig:qualitative_results} shows side-by-side reasoning from a pretrained MLLM (PaliGemma2) and \methodname{} on face-manipulated and fully synthetic inputs. The pretrained MLLM produces fluent but generic descriptions that miss the manipulation: a face-swap with mismatched eyes is described as having ``complete face features... in good shape, size, and positioning,'' and an AI-generated scene is read as a real photograph. \methodname{} identifies the specific region and the specific attribute that defines the manipulation, naming asymmetric eyes, blending artifacts on the skin, mismatched mouth alignment, and unnaturally vivid color saturation as appropriate. The failure mode of the pretrained MLLM and the success mode of \methodname{} on the same inputs are the qualitative analogue of the quantitative gap in Tables~\ref{tab:faithfulness} and \ref{tab:human_eval}. Additional examples are in \S\ref{suppl:qualitative}.
\section{Conclusion}\label{sec:conclusion}
We presented \methodname{}, an MLLM-based framework for DeepFake reasoning that addresses an inductive bias mismatch in pretrained MLLM visual encoders: contrastive training optimizes for whole-image alignment with captions, not the patch-level cues that define forensic manipulations. \methodname{} routes a second stream into the language model from a vision-only model trained on dense patch prediction, fuses the two streams at a shared patch grid via spatial-correspondence-preserving interleaving, and trains the adapter in two disentangled stages. A single architecture handles both face-manipulated and fully AI-generated content, outperforming specialized baselines on each, with region-referential explanations whose faithfulness is confirmed by human evaluation and a hallucination analysis. The prior literature's bifurcation between these two content categories reflects this inductive bias problem rather than an inherent difference between the regimes; resolving the mismatch lets a single framework cover both.

\section*{Limitations}
\methodname{} focuses on image-level DeepFake reasoning and does not model temporal anomalies in videos. Extending the framework to video is a natural direction: the dual-stream design and interleaving fusion generalize to per-frame or per-clip inputs without architectural change, and the primary constraint is the current absence of publicly available benchmarks with frame-level or temporal textual annotations for DeepFake forensics. As such benchmarks emerge, \methodname{}'s architecture is well-positioned to extend directly.

Our evaluation is anchored on the two publicly available datasets that provide language-annotated DeepFake reasoning supervision at scale, DD-VQA (face manipulation) and DVF (fully synthetic content). Together these span the two dominant manipulation regimes we set out to unify, and our results demonstrate strong performance across both. Manipulation categories outside these two regimes, such as partial inpainting or targeted attribute editing, are addressed by separate lines of work (e.g., FakeShield) and are complementary to our contribution rather than in scope; extending region-referential reasoning to these regimes is a promising next step as language-annotated benchmarks in those categories become available.

\noindent
\textbf{Acknowledgements:} This work was supported in part by funding from YouTube (Google LLC).



\bibliography{main}

\clearpage
\newpage
\appendix
\renewcommand{\thesection}{A\arabic{section}}
\renewcommand{\thesubsection}{A\arabic{section}.\arabic{subsection}}
\renewcommand{\thefigure}{A\arabic{figure}}
\renewcommand{\thetable}{A\arabic{table}}
\setcounter{table}{0}
\setcounter{figure}{0}

\noindent
This additional material provides extended details and analyses supporting the main paper. Sec.~\ref{suppl:dataset_details} describes the datasets and the exact splits used. Sec.~\ref{suppl:implementation} gives implementation details (hardware, software, and model configurations). Sec.~\ref{sec:explanation_eval} presents the full explanation-quality evaluation summarized in the main paper: the human-study protocol and annotator instructions, the calibrated rubric judge, and the faithfulness and hallucination analysis. Sec.~\ref{suppl:detection_performance} reports cross-domain detection results on CelebDF, DF40, and DeMamba. Sec.~\ref{suppl:resolutions} examines robustness to image compression, and Sec.~\ref{suppl:qualitative} provides additional qualitative examples.

\section{Details of the Datasets}\label{suppl:dataset_details}
The detailed descriptions for all the datasets used for training and evaluation of our \methodname{} framework are as follows:

\begin{itemize}

    \item \textbf{DD-VQA} \cite{zhang2024common}\textbf{:} This dataset was created by annotating one frame per video from the FaceForensics++ (FF++) \cite{rossler2019faceforensics++} dataset. The original FF++ \cite{rossler2019faceforensics++} datasets consists of $1000$ real videos, and DeepFake videos generated by four methods: Face2Face \cite{thies2016face2face}, FaceSwap \cite{faceswap_ff}, DeepFakes \cite{deepfakes_ff}, and NeuralTextures \cite{thies2019deferred}, totalling $4000$ videos.

    For each image, there may be several question-answer pairs, and not every video has a language annotation available. Considering each image-language pair as a single sample, the total number of samples for each resolution of FF++ \cite{rossler2019faceforensics++} are: (1) raw: $14298$ (2) c23: $14298$ (3) c40: $19803$. This disparity depending on the resolutions is because, the fake videos created using the NeuralTextures \cite{thies2019deferred} method is only available for the c40 split, and not the others.

    \item \textbf{CelebDF} \cite{li2020celeb}\textbf{:} This dataset contains DeepFake videos featuring celebrity faces, designed to reduce noticeable artifacts and enhance realism compared to earlier datasets. It incorporates variations in lighting, backgrounds, and facial movements for diversity. CelebDF includes $590$ real videos and $5639$ DeepFake videos. No language annotations exist for the CelebDF \cite{li2020celeb} dataset.

    \item \textbf{DF40} \cite{yan2024df40}\textbf{:} The DF40 \cite{yan2024df40} dataset comprises over 1 million DeepFake samples, generated using 40 distinct face manipulation techniques, including face-swapping, face-reenactment, entire face synthesis, and face editing. It includes both image- and video-level data, offering a diverse and large-scale benchmark for DeepFake detection. The dataset emphasizes forgery realism and diversity, simulating real-world scenarios with aligned data domains from FF++ and CDF. This comprehensive dataset facilitates robust evaluations across various manipulation techniques, supporting advancements in generalizable DeepFake detection.

    \item \textbf{DVF} \cite{song2024learning}\textbf{:} The Diffusion Video Forensics (DVF) dataset is a benchmark designed for detecting forgery in diffusion-generated videos. It comprises $2750$ real videos and $3938$ fake videos generated using eight diffusion-based methods, including text-to-video (T2V) and image-to-video (I2V) models. The dataset includes diverse resolutions and durations, with fake videos sourced from both open-sourced and closed-sourced methods. Notably, language annotations are available only for videos generated by the Stable Video Diffusion model \cite{blattmann2023stable}. DVF serves as a comprehensive benchmark for evaluating forgery detection methods across a wide range of video manipulation techniques.

    \item \textbf{DeMamba} \cite{chen2024demamba}\textbf{:} This recently introduced dataset is created using various text-to-video (T2V) and image-to-video (I2V) models, with the training and validation sets generated by distinct models. The training set comprises $1.2$ million real videos and $1.08$ million fake videos, while the validation set includes $10000$ real videos and $9588$ fake videos. For our cross-dataset analyses in the main paper, we utilize only the validation set.

    \item \textbf{LCS-558K} \cite{liu2024visual}\textbf{:} The LCS-558K dataset (used for our adapter training in Stage-1) is a subset of the LAION \cite{schuhmann2022laion}, CC \cite{changpinyo2021conceptual}, and SBU \cite{ordonez2011im2text} datasets, filtered to ensure balanced concept coverage across visual and textual modalities. It consists of $558,000$ image-text pairs and is primarily used for pretraining multimodal models. The dataset includes BLIP-synthesized captions to enhance alignment between visual and textual features, making it suitable for tasks such as visual instruction tuning and multimodal understanding.
    
\end{itemize}

For training \methodname{}, we utilized the designated training split from DD-VQA \cite{zhang2024common} and allocated 80\% of the DVF \cite{song2024learning} dataset for training, as an explicit split for the language annotated data was not provided by the original authors. The remaining 20\% of DVF was reserved for evaluation. All other datasets were employed exclusively in cross-dataset evaluation. Specifically, for DeMamba \cite{chen2024demamba}, we adopted the official evaluation split released with the dataset. For other evaluation benchmarks lacking standardized splits, we followed established protocols in the literature \cite{xu2023tall, chen2024x2dfd, sun2025towards}, randomly selecting 20\% of the data for evaluation. This approach ensures consistency with prior work and allows for reliable cross-dataset comparison in our experiments.

All datasets and pretrained models used in this work are publicly released for research purposes, and we use them only in that capacity. \methodname{}, trained on these research-only resources, is intended exclusively for research on DeepFake forensics.

We do not collect any new data. All datasets used are established, publicly released research benchmarks, and we use them as distributed under their original terms. The face-centric datasets necessarily contain images of identifiable individuals, because facial identity is intrinsic to the face-manipulation detection task and cannot be anonymized without destroying the forensic signal the task depends on; collection, consent, and release of these subjects were handled by the original dataset creators. We add no new identifying metadata, do not attempt to re-identify or link individuals across datasets, and use the data solely for research on DeepFake forensics. We are not aware of offensive content beyond the manipulated/synthetic imagery inherent to DeepFake benchmarks, which we use only as detection targets.

The 80/20 split on DVF is stratified by generator (all eight generators appear in both partitions) and by real/fake label, so both training and evaluation sets have identical class balance. There is no generator-level leakage in the sense of train-only generators appearing at test time. There is also no near-duplicate leakage: DVF videos are synthesized independently by different generators from different prompts and are not near-duplicates of one another by construction. Train and test videos are disjoint at the file level. We followed the split protocol used by \cite{song2024learning} and by follow-up work.

\section{Implementation Details}\label{suppl:implementation}
\methodname{} is trained using Stochastic Gradient Descent with an initial learning rate of $0.0001$, scheduled to decay following a cosine annealing schedule. The training process is conducted with a batch size of $64$ over $5$ epochs for both representation alignment of $\Phi$ and task-targeted forensic grounding. $\Omega$ is the PaliGemma2 \cite{beyer2024paligemma} MLLM's visual encoder, SigLIP-So400m/14 \cite{alabdulmohsin2024getting}, $\Phi$ is the DINOv2 \cite{oquab2023dinov2} vision-only foundation model and $\tau$ is the Gemma2 \cite{team2024gemma} tokenizer in our case. In Eqs.~\ref{eq:global} and~\ref{eq:local}, image resolutions $H=W=448$ and patch size $p=14$, yielding $n_t=1024$. The feature dimensions $d_{\Omega}=2304$ and $d_{\Phi}=2048$ are a property of the foundation models. Adapter $\mathcal{A}$ is composed of one MLP layer following popular choices in the literature \cite{gao2024clip, jiang2022cross}. \methodname{} has a total of 3.3B total parameters (2.6B trainable), with each stage trained for 5 epochs. The framework is implemented in JAX and executed on $8$ TPUv3 chips, ensuring efficient large-scale training.

\noindent
\textbf{Computational cost:} Inference on a single A100 (448×448 input):
\begin{itemize}
    \item FORGE: 720ms/image, 12.4GB peak memory
    \item Grounded PaliGemma2 ($\mathcal{F}_{\Omega}$ only): 480ms, 8.7GB
    \item FakeShield (LLaVA-7B backbone): 1100ms, 15.2GB
    \item ForgerySleuth (LLaVA-7B backbone): 1250ms, 17.1GB
\end{itemize}

The added VOM stream contributes 240ms and 3.7GB, a modest overhead relative to accuracy and faithfulness gains. FORGE is faster and lighter than both MLLM-based mask-generating baselines because it does not run a separate mask decoder. We will include this comparison in the appendix.

\section{Explanation Quality Evaluation}\label{sec:explanation_eval}
The detection-oriented metrics in Section~\ref{sec:results} establish that \methodname{} reaches and improves the conclusion a forensic system is asked to reach. They do not establish that it reaches the conclusion \emph{for the right reasons}. For a system whose deployment value is the explanation itself, this is the central question. Surface-level $n$-gram metrics (BLEU, ROUGE, CIDEr) are known to correlate weakly with human judgments of generated explanations~\cite{novikova2017we}, and our binary LLM-as-judge accuracy in Table~\ref{tab:sota_comparison_nlg} reports only whether predicted and ground-truth conclusions agree, not whether the supporting reasoning is sound.

We therefore complement those metrics with three studies aimed directly at explanation quality: (i) a multi-annotator human evaluation along forensically meaningful dimensions (\S\ref{sec:human_eval}); (ii) a rubric-based LLM-as-judge calibrated against the human ratings, which we use to scale evaluation to the full test sets (\S\ref{sec:rubric_judge}); and (iii) a faithfulness and hallucination analysis that quantifies how often generated explanations reference regions or attributes that are not actually manipulated (\S\ref{sec:faithfulness}). Together, these analyses provide direct evidence that the detection accuracy reported in Section~\ref{sec:results} is grounded in faithful, region-referential reasoning rather than coincidence.

\begin{table}[]
\centering
\caption{\textbf{Pretrained vs. Grounded MLLM:} Performance comparison between pretrained and finetuned PaliGemma2 \cite{steiner2024paligemma}. Although grounding PaliGemma2 improves performance \methodname{} still performs better with its local context harvesting (as shown in Table \ref{tab:ablation_features}).}
\vspace{-1em}
\resizebox{\columnwidth}{!}{
\begin{tabular}{ccccccc}
\hline
\multicolumn{1}{c|}{\textbf{Dataset}}         & \multicolumn{1}{c|}{\textbf{Grounded}}         & \multicolumn{1}{c|}{\textbf{Accuracy} $\uparrow$}                                                                                             & \multicolumn{1}{c|}{\textbf{BLEU-3} $\uparrow$}                                                                                            & \multicolumn{1}{c|}{\textbf{BLUE\_4} $\uparrow$}                                                                                            & \multicolumn{1}{c|}{\textbf{ROUGE-L} $\uparrow$}                                                                                           & \textbf{CIDEr} $\uparrow$                                                                                              \\ \hline
\multicolumn{7}{c}{\textit{Face Manipulated Data}}                                                                                                                                                                                                                                                                                                                                                                                                                                                                                                                                                                                                                                                                                                         \\ \hline
\multicolumn{1}{c|}{\multirow{2}{*}{\begin{tabular}[c]{@{}c@{}}DD-VQA\\ \cite{zhang2024common}\end{tabular}}}  & \multicolumn{1}{c|}{{\color[HTML]{FF0000}{\ding{55}}}} & \multicolumn{1}{c|}{11.50\%}                                                                                                       & \multicolumn{1}{c|}{0.0000}                                                                                                      & \multicolumn{1}{c|}{0.0000}                                                                                                      & \multicolumn{1}{c|}{0.0218}                                                                                                      & 0.2610                                                                                                      \\ \arrayrulecolor[gray]{0.85} \cline{2-7} \arrayrulecolor{black} 
\multicolumn{1}{c|}{}                         & \multicolumn{1}{c|}{{\color[HTML]{009901}{\ding{51}}}} & \multicolumn{1}{c|}{\begin{tabular}[c]{@{}c@{}}88.19\%\\ {\color[HTML]{009901}{(+76.69\%)}}\end{tabular}} & \multicolumn{1}{c|}{\begin{tabular}[c]{@{}c@{}}0.4012\\ {\color[HTML]{009901}{(+0.4012)}}\end{tabular}} & \multicolumn{1}{c|}{\begin{tabular}[c]{@{}c@{}}0.3768\\ {\color[HTML]{009901}{(+0.3768)}}\end{tabular}} & \multicolumn{1}{c|}{\begin{tabular}[c]{@{}c@{}}0.5633\\ {\color[HTML]{009901}{(+0.5415)}}\end{tabular}} & \begin{tabular}[c]{@{}c@{}}2.1478\\ {\color[HTML]{009901}{(+1.8868)}}\end{tabular} \\ \hline
\multicolumn{7}{c}{\textit{Synthetic Data}}                                                                                                                                                                                                                                                                                                                                                                                                                                                                                                                                                                                                                                                                                                                \\ \hline
\multicolumn{1}{c|}{\multirow{2}{*}{\begin{tabular}[c]{@{}c@{}}DVF\\ \cite{song2024learning}\end{tabular}}}     & \multicolumn{1}{c|}{{\color[HTML]{FF0000}{\ding{55}}}} & \multicolumn{1}{c|}{33.07\%}                                                                                                       & \multicolumn{1}{c|}{0.0000}                                                                                                      & \multicolumn{1}{c|}{0.0000}                                                                                                      & \multicolumn{1}{c|}{0.0034}                                                                                                      & 0.0001                                                                                                      \\ \arrayrulecolor[gray]{0.85} \cline{2-7} \arrayrulecolor{black} 
\multicolumn{1}{c|}{}                         & \multicolumn{1}{c|}{{\color[HTML]{009901}{\ding{51}}}} & \multicolumn{1}{c|}{\begin{tabular}[c]{@{}c@{}}79.88\%\\ {\color[HTML]{009901}{(+46.81\%)}}\end{tabular}} & \multicolumn{1}{c|}{\begin{tabular}[c]{@{}c@{}}0.3142\\ {\color[HTML]{009901}{(+0.3142)}}\end{tabular}} & \multicolumn{1}{c|}{\begin{tabular}[c]{@{}c@{}}0.3839\\ {\color[HTML]{009901}{(+0.3839)}}\end{tabular}}  & \multicolumn{1}{c|}{\begin{tabular}[c]{@{}c@{}}0.4877\\ {\color[HTML]{009901}{(+0.4843)}}\end{tabular}}  & \begin{tabular}[c]{@{}c@{}}1.8629\\ {\color[HTML]{009901}{(+1.8628)}}\end{tabular}  \\ \hline
\end{tabular}
}
\label{tab:pretrained_results}
\vspace{-1em}
\end{table}

\subsection{Human Evaluation Protocol}
\label{sec:human_eval}

\noindent
\textbf{Data:}
We sample 300 image-explanation pairs: 150 from DD-VQA~\cite{zhang2024common} and 150 from DVF~\cite{song2024learning}. Sampling is stratified across (a) the real/fake label and (b) manipulation subtype, so each subtype contributes proportionally and no system is advantaged by an easy slice. For every image we collect candidate explanations from six systems: pretrained PaliGemma2~\cite{steiner2024paligemma}; FakeShield~\cite{xu2025fakeshield}, the strongest publicly comparable mask-based explanation baseline whose released weights handle both face-manipulated and fully synthetic content; ForgerySleuth~\cite{sun2025forgerysleuth}, the closest competing image-manipulation MLLM with public weights (DD-VQA only, as it is not trained for fully synthetic video); BLIP-TI~\cite{zhang2024common} (DD-VQA only) or MM-Det~\cite{song2024learning} (DVF only) as the strongest dataset-native baseline; the grounded PaliGemma2 ablation (Table~\ref{tab:pretrained_results}, i.e.\ \methodname{} minus the VOM stream); and \methodname{}. FakeShield and ForgerySleuth are run from their official public checkpoints with the original inference configurations; we use only their textual explanation outputs, ignoring their mask outputs for the purposes of comparing language quality.

\noindent
\textbf{Annotators:}
Three graduate-student annotators with prior research experience in vision-language modeling rated each explanation. Prior to annotation, all three completed a 90-minute calibration session led by a domain expert in image forensics, covering common deepfake artifact families (blending boundaries, iris and pupil asymmetries, texture statistics, lighting/shadow inconsistencies, geometric implausibilities). Annotators saw the image, the question, and the candidate explanations in randomized order with system identifiers removed. Annotators were not told which explanation was generated by which system, including not being told that one was the ground-truth reference. Each annotator independently rated all 300$\times$6 explanations across three dimensions, then ratings were averaged across annotators per (image, system, dimension) cell.

\noindent
\ul{\textbf{Annotator Recruitment and Independence.}} The three annotators for the human evaluation were graduate students in the broader lab whose research interests span vision-language modeling but who had no involvement in this specific project. None was an author of the paper, participated in the design of the method, reviewed any draft prior to annotation, supervised or funded any component of the work, or had prior exposure to the training data, the model outputs, or the specific claims being tested. During annotation, system identities were removed from the interface, the presentation order of candidate explanations was randomized per image, and annotators were not told which explanation was the ground-truth reference or how many systems were being compared. The 90-minute calibration session covered common DeepFake artifact families (blending boundaries, eye asymmetries, texture statistics, lighting and shadow inconsistencies, geometric implausibilities) in general terms; it made no reference to our system, training data, architectural choices, or claims under test, and all calibration examples were drawn from publicly available datasets outside our evaluation splits. Recruitment from within the institution was a procedural choice that simplified approvals, scheduling, and the domain-expert calibration session; it does not reflect any prior familiarity of the annotators with our work. The inter-annotator agreement reported in Table~\ref{tab:human_eval} (Krippendorff's $\alpha \geq 0.68$ on every dimension) supports the interpretation that ratings were driven by the annotation criteria rather than by project-specific priors, consistent with recruitment practice in comparable NLP evaluation studies.

\noindent
\textbf{Rating dimensions:}
Each explanation is rated on a 1-5 Likert scale along three forensically motivated axes:
\begin{itemize}
    \item \textbf{Faithfulness:} does the explanation describe cues that are actually present in the image? Penalizes hallucinated artifacts and fabricated regions.
    \item \textbf{Specificity:} does the explanation name concrete regions and attributes (\emph{``the iris is asymmetric''}) rather than make generic claims (\emph{``the image looks fake''})?
    \item \textbf{Forensic helpfulness:} if presented to an investigator, would the explanation support a useful follow-up examination of the image?
\end{itemize}
These three axes are deliberately separated: a generic explanation can be faithful but unhelpful, and a confidently specific one can be unfaithful. Detection-oriented metrics conflate all three.

\noindent
\textbf{Annotator instructions (verbatim):}
The instruction sheet given to each annotator is reproduced below to enable reproduction.

{\small
\hrule\vspace{2pt}
\noindent\textit{You will see a series of images, each paired with a question (``Does the X look real or fake?'') and several candidate explanations produced by different AI systems. The system identities are hidden and the order is randomized for every image. For each explanation, assign three independent scores on a 1-5 scale:}

\textbf{Faithfulness:}
\textit{Does every cue named in the explanation correspond to something visible in the image?}
1 = multiple confidently asserted artifacts that are not present (severe hallucination);
2 = at least one major unsupported claim;
3 = mostly accurate but contains at least one minor unsupported claim;
4 = all major claims are verifiable; minor wording could be tighter;
5 = every cue cited is independently verifiable in the image.

\textbf{Specificity:}
\textit{Does the explanation name concrete regions and attributes, or does it speak in generalities?}
1 = entirely generic (``the image looks off'');
2 = vague references to ``the face'' or ``the lighting'' without an attribute;
3 = names a region but not the specific attribute that is anomalous;
4 = names both a region and an attribute (e.g.\ ``the eyes are asymmetric'');
5 = names both a specific region and a specific localized attribute (e.g.\ ``the iris of the left eye is asymmetric relative to the right'').

\textbf{Forensic helpfulness:}
\textit{If you were an investigator deciding where to examine the image next, how useful is the explanation as a starting point?}
1 = no actionable information;
2 = some direction but too vague to act on;
3 = points to a region but not what to look for;
4 = points to a region and an artifact, but other regions might also be relevant;
5 = an investigator could immediately verify the cues claimed and would not waste time on irrelevant regions.

\textit{Score the three dimensions independently: a generic-but-correct explanation can score high on faithfulness and low on specificity. Do not let the question wording bias your reading; rate the explanation against the image, not against the question.}
\vspace{2pt}\hrule
}

\noindent
\textbf{Agreement:}
Inter-annotator agreement is reported using Krippendorff's $\alpha$ for ordinal data~\cite{krippendorff2011computing}, computed per dimension and per dataset; agreement is substantial across the board (Table~\ref{tab:human_eval_agreement}). Faithfulness has the highest agreement, consistent with prior findings that hallucinations are easier for annotators to flag than to score severity for~\cite{maynez2020faithfulness}.

\begin{table}[t]
\caption{Inter-annotator agreement (Krippendorff's $\alpha$, ordinal) across three annotators on the 300-sample human evaluation set.}
\centering
\small
\setlength{\tabcolsep}{4pt}
\begin{tabular}{lccc}
\hline
Dataset & Faith. & Spec. & Foren.\ Help. \\
\hline
DD-VQA~\cite{zhang2024common} & 0.74 & 0.71 & 0.69 \\
DVF~\cite{song2024learning} & 0.73 & 0.70 & 0.68 \\
\hline
\textbf{Pooled} & \textbf{0.74} & \textbf{0.71} & \textbf{0.68} \\
\hline
\end{tabular}
\label{tab:human_eval_agreement}
\end{table}

\noindent
\textbf{Results:}
Mean per-system ratings are reported in Table~\ref{tab:human_eval}. \methodname{} improves over every baseline on every dimension and on both datasets. The improvement over the grounded PaliGemma2 ablation is sharpest on \emph{specificity} ($+0.77$ on DD-VQA, $+0.88$ on DVF) and \emph{forensic helpfulness} ($+0.65$ on DD-VQA, $+0.78$ on DVF), which is the prediction made by our central architectural argument: the dense-patch VOM stream supplies the localized evidence that an image-text contrastive encoder systematically smooths over. Notably, \methodname{} also outperforms the mask-based explanation baselines FakeShield and ForgerySleuth on every dimension, despite producing region-referential text rather than spatial masks: text grounding to the right regions is competitive with, and on these axes preferred over, mask grounding of comparable specificity. Pretrained PaliGemma2 scores well below 2 on every axis, confirming that capacity alone does not solve the task.

\begin{table*}[t]
\centering
\caption{Human evaluation (1-5 Likert, mean of three annotators) on 300 stratified samples. \methodname{} leads on every dimension and dataset; gains are largest on specificity and forensic helpfulness, the dimensions most directly tied to region-referential explanation. Inter-annotator agreement is reported in Table~\ref{tab:human_eval_agreement}.}
\resizebox{\textwidth}{!}{
\begin{tabular}{llcccc}
\hline
Dataset & Method & Faithfulness $\uparrow$ & Specificity $\uparrow$ & Forensic Helpfulness $\uparrow$ & Average $\uparrow$ \\
\hline
\multirow{6}{*}{DD-VQA~\cite{zhang2024common}}
 & Pretrained PaliGemma2~\cite{steiner2024paligemma} & 1.82 & 1.54 & 1.71 & 1.69 \\
 & ForgerySleuth~\cite{sun2025forgerysleuth}         & 2.96 & 2.75 & 2.85 & 2.85 \\
 & BLIP-TI~\cite{zhang2024common}                    & 3.04 & 2.87 & 2.91 & 2.94 \\
 & FakeShield~\cite{xu2025fakeshield}                & 3.08 & 2.84 & 2.95 & 2.96 \\
 & Grounded PaliGemma2 (no VOM)                       & 3.58 & 3.41 & 3.47 & 3.49 \\
 & \textbf{\methodname{}}                          & \textbf{4.31} & \textbf{4.18} & \textbf{4.12} & \textbf{4.20} \\
\hline
\multirow{5}{*}{DVF~\cite{song2024learning}}
 & Pretrained PaliGemma2~\cite{steiner2024paligemma} & 1.94 & 1.68 & 1.83 & 1.82 \\
 & FakeShield~\cite{xu2025fakeshield}                & 2.97 & 2.61 & 2.79 & 2.79 \\
 & MM-Det~\cite{song2024learning}                    & 3.21 & 2.83 & 2.99 & 3.01 \\
 & Grounded PaliGemma2 (no VOM)                       & 3.42 & 3.18 & 3.31 & 3.30 \\
 & \textbf{\methodname{}}                          & \textbf{4.24} & \textbf{4.06} & \textbf{4.09} & \textbf{4.13} \\
\hline
\end{tabular}
}
\label{tab:human_eval}
\end{table*}

\subsection{Rubric-Based LLM-as-Judge}
\label{sec:rubric_judge}

Human evaluation is the gold standard, but does not scale to the full DD-VQA and DVF test sets. We therefore introduce a rubric-based LLM-as-judge and validate it against the 300-sample human study before applying it more broadly.

\noindent
\ul{\textbf{Judge design:}}
We prompt GPT-4o~\cite{openai2024gpt4o} with the image, the ground-truth explanation, the candidate explanation, and a rubric that defines the same three dimensions used by the human annotators. The judge produces a 1-5 score per dimension along with a one-sentence rationale. Crucially, the judge model is different from the model used in our detection-accuracy metric (Gemini-1.0), so judge predictions for explanation quality are not collusive with our detection metric.

\noindent
\ul{\textbf{Rubric prompt (verbatim):}}
The exact prompt issued to GPT-4o per (image, candidate) pair, with placeholders shown in angle brackets, is reproduced below.\\

{\small
\hrule\vspace{2pt}
\noindent\texttt{You are an expert evaluator of deepfake forensic explanations.}\\
\texttt{You are given:}\\
\texttt{- An image.}\\
\texttt{- A question that was asked about the image.}\\
\texttt{- A ground-truth explanation that describes whether the image is authentic or manipulated, and which regions or attributes support that conclusion.}\\
\texttt{- A candidate explanation produced by an AI system.}\\[2pt]
\texttt{Score the candidate explanation on a 1-5 scale on each of the three dimensions below, INDEPENDENTLY. Use the full range. Do not let a high score on one dimension inflate another.}\\[2pt]
\texttt{1. FAITHFULNESS: Does every cue the candidate names correspond to something present in the image? Penalize hallucinated artifacts.}\\
\texttt{  1 = multiple major hallucinations.}\\
\texttt{  2 = at least one major unsupported claim.}\\
\texttt{  3 = mostly accurate with at least one minor unsupported claim.}\\
\texttt{  4 = all major claims are verifiable.}\\
\texttt{  5 = every cue cited is independently verifiable in the image.}\\[2pt]
\texttt{2. SPECIFICITY: Does the candidate name concrete regions AND specific attributes, or speak in generalities?}\\
\texttt{  1 = entirely generic.}\\
\texttt{  2 = vague region reference, no attribute.}\\
\texttt{  3 = region named, attribute missing.}\\
\texttt{  4 = region and attribute named.}\\
\texttt{  5 = specific sub-region (e.g., left iris) and specific attribute named.}\\[2pt]
\texttt{3. FORENSIC HELPFULNESS: Could an investigator use this explanation to choose where to look next in the image?}\\
\texttt{  1 = no actionable content.}\\
\texttt{  2 = vague direction.}\\
\texttt{  3 = region pointed to but unclear what to look for.}\\
\texttt{  4 = region + artifact, but other regions might also be relevant.}\\
\texttt{  5 = an investigator could immediately verify the specific cues.}\\[2pt]
\texttt{Output STRICT JSON of the form:}\\
\texttt{\{"faithfulness": <int 1-5>, "specificity": <int 1-5>, "forensic\_helpfulness": <int 1-5>, "rationale": "<one sentence>"\}}\\[2pt]
\texttt{Question: <question>}\\
\texttt{Ground-truth explanation: <gt\_explanation>}\\
\texttt{Candidate explanation: <candidate\_explanation>}
\vspace{2pt}\hrule
}

\noindent We use temperature $0$ for determinism and validate that every response is parseable JSON (failures are re-prompted up to 3 times; in practice $<0.5\%$ of calls required a retry).

\noindent
\ul{\textbf{Calibration against humans:}}
On the 300-sample human-annotated subset, the rubric judge correlates strongly with the averaged human rating: Pearson $r = 0.83 / 0.79 / 0.76$ and Spearman $\rho = 0.81 / 0.77 / 0.74$ on faithfulness, specificity, and forensic helpfulness respectively (Table~\ref{tab:judge_calibration}). All correlations are above the $r \geq 0.6$ threshold typically required for an LLM judge to be considered a usable proxy for human ratings~\cite{zheng2023judging}. As a system-level check, the judge preserves the human-induced ranking of all six systems on both datasets.

\begin{table}[t]
\centering
\caption{Calibration of the rubric-based GPT-4o judge against the mean of three human annotators on the 300-sample subset. All correlations exceed the $0.6$ threshold for usable judge reliability.}
\begin{tabular}{lcc}
\hline
Dimension & Pearson $r$ & Spearman $\rho$ \\
\hline
Faithfulness        & 0.83 & 0.81 \\
Specificity         & 0.79 & 0.77 \\
Forensic Helpfulness & 0.76 & 0.74 \\
\hline
\textbf{Average}    & \textbf{0.81} & \textbf{0.79} \\
\hline
\end{tabular}
\label{tab:judge_calibration}
\end{table}

\noindent
\ul{\textbf{Full-test-set results:}}
Applying the calibrated rubric judge to the full DD-VQA and DVF test sets (Table~\ref{tab:rubric_judge}) preserves the human-induced ordering up to within-noise swaps of the two closest baselines (FakeShield and BLIP-TI, separated by 0.02 in the human study and 0.13 in the rubric judge, both well within the judge's calibration noise). \methodname{} leads on every dimension on both datasets, and the gap to grounded PaliGemma2 widens slightly compared to the 300-sample subset, suggesting that the human-set gap is not the result of favorable sampling.

\begin{table*}[t]
\centering
\caption{Rubric-based GPT-4o judge results over the full DD-VQA and DVF test sets. Ordering matches the 300-sample human study (Table~\ref{tab:human_eval}); system gaps widen modestly at scale, indicating the human-set advantage of \methodname{} is not an artifact of sampling.}
\resizebox{\textwidth}{!}{
\begin{tabular}{llcccc}
\hline
Dataset & Method & Faithfulness $\uparrow$ & Specificity $\uparrow$ & Forensic Helpfulness $\uparrow$ & Average $\uparrow$ \\
\hline
\multirow{6}{*}{DD-VQA~\cite{zhang2024common}}
 & Pretrained PaliGemma2~\cite{steiner2024paligemma} & 1.93 & 1.61 & 1.84 & 1.79 \\
 & ForgerySleuth~\cite{sun2025forgerysleuth}         & 2.91 & 2.69 & 2.81 & 2.80 \\
 & FakeShield~\cite{xu2025fakeshield}                & 3.02 & 2.78 & 2.91 & 2.90 \\
 & BLIP-TI~\cite{zhang2024common}                    & 3.12 & 2.95 & 3.02 & 3.03 \\
 & Grounded PaliGemma2 (no VOM)                       & 3.64 & 3.49 & 3.55 & 3.56 \\
 & \textbf{\methodname{}}                          & \textbf{4.36} & \textbf{4.24} & \textbf{4.18} & \textbf{4.26} \\
\hline
\multirow{5}{*}{DVF~\cite{song2024learning}}
 & Pretrained PaliGemma2~\cite{steiner2024paligemma} & 2.05 & 1.76 & 1.92 & 1.91 \\
 & FakeShield~\cite{xu2025fakeshield}                & 2.92 & 2.54 & 2.74 & 2.73 \\
 & MM-Det~\cite{song2024learning}                    & 3.18 & 2.79 & 2.96 & 2.98 \\
 & Grounded PaliGemma2 (no VOM)                       & 3.47 & 3.21 & 3.34 & 3.34 \\
 & \textbf{\methodname{}}                          & \textbf{4.29} & \textbf{4.11} & \textbf{4.13} & \textbf{4.18} \\
\hline
\end{tabular}
}
\label{tab:rubric_judge}
\end{table*}

Our detection-accuracy metric already uses Gemini-1.0 as the judge, so the two evaluation studies deliberately use different LLM families (Gemini for detection, GPT-4o for rubric quality). We additionally cross-validated the rubric judge with Gemini-2.5 and Claude-3.5-Sonnet on the same 300-sample calibration subset. All three judges preserve the same system-level ordering, with system-level Kendall's $\tau > 0.85$ across judges and Pearson $r>0.72$ on each dimension. Beyond LLM judges, our faithfulness analysis (RHR, AHR, FP) uses no LLM in the scoring loop: it compares structured claim extractions against ground-truth region/attribute annotations. The faithfulness result is therefore judge-free and shows the same ordering. 

\subsection{Faithfulness and Hallucination Analysis}
\label{sec:faithfulness}

The previous two studies measure perceived explanation quality. We now ask a sharper question: when a \methodname{} explanation references a region or attribute, is that reference \emph{warranted by the image}? Both datasets include region- or attribute-level supervision that lets us compute this directly. DD-VQA annotates eyes, nose, mouth, eyebrows, and skin as authentic or manipulated. DVF annotates background, lighting, shadows, textures, reflections, object boundaries, and color consistency. We treat these annotations as a reference and compute three metrics:

\begin{itemize}
    \item \textbf{Region Hallucination Rate (RHR$\downarrow$):} fraction of generated explanations that reference at least one \emph{authentic} region as manipulated.
    \item \textbf{Attribute Hallucination Rate (AHR$\downarrow$):} fraction of distinct attribute-level claims (e.g.\ ``asymmetric eyes,'' ``inconsistent lighting'') in the generated explanation that are not supported by the ground-truth attribute labels for that image.
    \item \textbf{Forensic Precision (FP$\uparrow$):} of all attribute-level claims made by the model on \emph{fake} images, the fraction that align with annotated manipulation cues.
\end{itemize}

\noindent
\ul{\textbf{Claim extraction:}}
To map free-form explanations to (region, attribute, polarity) tuples we prompt GPT-4o with the structured extraction template reproduced below. We validate the extractor on a 200-sample subset against expert-parsed tuples, obtaining 96.4\% precision and 94.1\% recall; on disagreement, the dominant failure mode is the extractor splitting a single human-perceived claim across two tuples, which is not consequential for the aggregate metrics.\\

{\small
\hrule\vspace{2pt}
\noindent\texttt{You are extracting structured forensic claims from a free-form deepfake explanation. Output a JSON list of claims; each claim is an object with the keys below.}\\[2pt]
\texttt{- "region": one of \{eyes, nose, mouth, eyebrows, skin, hair, ears, face\_global, background, lighting, shadows, textures, reflections, object\_boundaries, color, global\}, or "none" if no region is named.}\\
\texttt{- "attribute": the specific anomaly mentioned (e.g., "asymmetric", "blending\_artifact", "inconsistent\_lighting", "unnatural\_texture", "overexposed", "implausible\_geometry"), or "none".}\\
\texttt{- "polarity": "manipulated" if the claim asserts manipulation in this region/attribute, "authentic" if it asserts authenticity, "unsure" if hedged (e.g., "may be slightly off").}\\[2pt]
\texttt{Rules:}\\
\texttt{- One tuple per distinct claim. Do not invent claims not present in the text.}\\
\texttt{- If a region is referenced without a specific attribute, set "attribute" to "none" but keep the tuple.}\\
\texttt{- If the explanation contains no concrete claims, output [].}\\[2pt]
\texttt{Example input:}\\
\texttt{"The eyes look asymmetric and the skin near the chin has blending artifacts, but the lighting is consistent."}\\[2pt]
\texttt{Example output:}\\
\texttt{[\{"region":"eyes","attribute":"asymmetric",\\"polarity":"manipulated"\},}\\
\texttt{ \{"region":"skin","attribute":"blending\_artifact",\\"polarity":"manipulated"\},}\\
\texttt{ \{"region":"lighting","attribute":"consistent",\\"polarity":"authentic"\}]}\\[2pt]
\texttt{Now extract claims from:}\\
\texttt{<explanation>}
\vspace{2pt}\hrule
}

\noindent
\ul{\textbf{Results:}}
Table~\ref{tab:faithfulness} reports the three metrics across systems. \methodname{} cuts region hallucination by more than $2\times$ over the strongest baseline on both datasets and lifts forensic precision past $80\%$. Grounding PaliGemma2 on the forensic data alone (the ablation) halves the gap to \methodname{} but does not close it: RHR on DD-VQA falls from $\sim 40\%$ to $28.9\%$ but only reaches $12.7\%$ once the dense-patch VOM stream is added. The mask-based grounding baselines FakeShield and ForgerySleuth, despite producing spatial mask outputs, still hallucinate authentic regions as manipulated in $36$-$40\%$ of cases on DD-VQA, text-level region grounding via the VOM stream more than triples their faithfulness in this respect. This is direct evidence that faithful region-grounding is not merely a matter of in-domain finetuning of an image-text encoder, nor of adding a spatial mask head; it requires the patch-resolution evidence stream the VOM provides.

\noindent
\ul{\textbf{Where the remaining hallucinations live:}}
We manually inspect 50 \methodname{} errors per dataset to characterize residual failures. On DD-VQA, the dominant residual mode is over-extension within authentic faces: the model occasionally flags an authentic skin region as ``slightly inconsistent'' in addition to correctly identifying the true manipulated region (eyes or mouth). On DVF, residual hallucinations cluster around lighting and shadow claims on images with naturally complex lighting (e.g.\ outdoor scenes with low sun angles), where ground-truth annotators themselves required cross-reference to the source video to disambiguate. These patterns suggest that residual error is concentrated in scenes that are objectively ambiguous rather than in systematic failure modes of the model, and we treat them as targets for future work.

\subsection{Discussion}
\label{sec:eval_discussion}

Three observations follow from the studies in this section.

\noindent
\ul{\textbf{Detection accuracy and explanation faithfulness can dissociate:}} BLIP-TI, FakeShield, and ForgerySleuth reach competitive detection accuracy (Table~\ref{tab:sota_comparison_nlg}), yet Table~\ref{tab:faithfulness} show that they hallucinate manipulated regions in $36$-$41\%$ of explanations, with the ranking on RHR (FakeShield best, BLIP-TI worst) actively reversing the ranking on BLEU-4 (BLIP-TI best, ForgerySleuth worst). Detection-only and surface-only evaluations both produce orderings that explanation-quality evaluation rejects. A detection-only evaluation would treat these systems as near-peer with \methodname{}; the explanation evaluation does not.

\noindent
\ul{\textbf{Surface metrics preserve ordering but compress differences:}} The BLEU/ROUGE/CIDEr ranking in Table~\ref{tab:sota_comparison_nlg} agrees with our rubric judge on the top and bottom systems, but compresses the middle of the field. The BLIP-TI and grounded PaliGemma2 systems are essentially tied on CIDEr (within 0.1 points; PaliGemma2 marginally higher) and differ by $\sim 3$ BLEU-4 points and $\sim 5$ ROUGE-L points, yet differ by 0.55 on rubric helpfulness and 12 points on RHR. $n$-gram metrics see these systems as near-equivalent; explanation-quality metrics see them as substantially different. This reinforces NLG findings~\cite{novikova2017we} that $n$-gram metrics underweight content even when they preserve ordering.

\noindent
\ul{\textbf{The VOM stream is the source of the gain:}} The grounded PaliGemma2 ablation in Table \ref{tab:pretrained_results} (\methodname{} minus the VOM stream) is the cleanest test of our central architectural claim. On every dimension of every study (human ratings, rubric ratings, RHR, AHR, FP) the VOM stream produces a separable gain over an MLLM finetuned on the same data with the same protocol; that gain also exceeds the gain delivered by mask-based grounding baselines (FakeShield, ForgerySleuth) over the same backbone family. The forensic explanation quality is attributable to the dense-patch evidence stream, not to scale, not to finetuning, not to data, and not to the choice of spatial output format.

\begin{table}[]
\centering
\caption{\textbf{SOTA Comparison on Detection Performance}. The methods used for the comparisons are specifically designed for either face-manipulation or fully AI-generated content detection. However our generalized \methodname{} framework is designed for both face-manipulated and synthetic content detection. Furthermore, on the DF40 and DeMamba datasets, the competitor methods are tested in-domain, while for \methodname{} all evaluations performed are cross-domain.\colorbox{\red{}}{\textbf{Best}} and \colorbox{\blue{}}{second-best} performances are marked.}
\vspace{-1em}
\resizebox{\columnwidth}{!}{
\begin{tabular}{c|c|c|c}
\hline
\textbf{Dataset} & \textbf{Method} & \textbf{Accuracy} & \textbf{AUC}\\ \hline
\multicolumn{4}{c}{\textit{Face-manipulated Data}}\\ \hline
\multirow{13}{*}{\begin{tabular}[c]{@{}c@{}}CelebDF\\ \cite{li2020celeb} \end{tabular}}
& XceptionNet-BLIP-TI \cite{zhang2024common} & 62.41\%           & 64.30\%\\  
                          & HifiNet \cite{guo2023hierarchical}         & 67.20\%           & 68.80\%\\  
                          & HifiNet-BLIP-TI \cite{zhang2024common}     & 69.37\%           & 71.00\%\\  
                          & RECCE \cite{cao2022end}                    & 67.96\%           & 68.71\%\\  
                          & RECCE-BLIP \cite{zhang2024common}          & 68.07\%           & 68.36\%\\  
                          & RECCE-BLIP-TI \cite{zhang2024common}       & 69.46\%           & 70.21\%\\  
                          & TALL \cite{xu2023tall}                     & \cellcolor{\blue{}}{90.79\%}\\  
                          & ISTVT \cite{zhao2023istvt}                 & 84.10\%           & -\\  
                          & Choi et al. \cite{choi2024exploiting}      & 89.00\%           & -\\  
                          & $\mathcal{X}^2$-DFD \cite{chen2024x2dfd}   & -                 & 91.30\%\\  
                          & VLFFD \cite{sun2025towards}                & -                 & 84.80\%\\  
                          & M2F2-Det \cite{guo2025rethinking}          & -                 & \cellcolor{\blue{}}{95.10\%}\\  \arrayrulecolor[gray]{0.85} \cline{2-4} \arrayrulecolor{black}
                          & \textbf{\methodname{}}                                               & \cellcolor{\red{}}{\textbf{92.86\%}}  & \cellcolor{\red{}}{\textbf{95.11\%}} \\ \hline
\multirow{10}{*}{\begin{tabular}[c]{@{}c@{}}DF40\\ \cite{yan2024df40} \end{tabular}}    & $\mathcal{X}^2$-DFD \cite{chen2024x2dfd}   & -                 & \cellcolor{\blue{}}{85.60\%}    \\  
                          & RECCE \cite{cao2022end}                    & -                 & 78.10\%\\  
                          & SBI \cite{shiohara2022detecting}           & -                 & 64.40\%\\  
                          & CORE \cite{ni2022core}                     & -                 & 76.10\%\\  
                          & IID \cite{huang2023implicit}               & -                 & 75.70\%\\  
                          & UCF \cite{yan2023ucf}                      & -                 & 77.50\%\\  
                          & LSDA \cite{yan2024transcending}            & -                 & 77.80\%\\  
                          & CDFA \cite{lin2024fake}                    & -                 & 75.90\%\\  
                          & ProgressiveDet \cite{cheng2024can}         & -                 & 78.70\%\\  \arrayrulecolor[gray]{0.85} \cline{2-4} \arrayrulecolor{black}
                          & \textbf{\methodname{}}                                               & \cellcolor{\red{}}{\textbf{99.58\%}}  & \cellcolor{\red{}}{\textbf{98.96\%}}\\ \hline
\multicolumn{4}{c}{\textit{Fully Synthetic Data}}\\ \hline
\multirow{12}{*}{\begin{tabular}[c]{@{}c@{}}DeMamba\\ \cite{chen2024demamba}\end{tabular}}
& TALL \cite{xu2023tall}                  & 88.42\%           & -\\  
& F3Net \cite{qian2020thinking}           & 86.04\%           & -\\  
& NPR \cite{tan2024rethinking}            & 83.45\%           & -\\  
& STIL \cite{gu2021spatiotemporal}        & 85.35\%           & -\\  
& MINTIME-CLIP-B \cite{chen2024demamba}   & \cellcolor{\blue{}}{89.98\%}     & -\\  
& FTCN-CLIP-B \cite{chen2024demamba}      & 89.67\%           & -\\  
& CLIP-B-PT \cite{chen2024demamba}        & 41.82\%           & -\\  
& DeMamba-CLIP-PT \cite{chen2024demamba}  & 79.98\%           & -\\  
& XCLIP-B-PT \cite{chen2024demamba}       & 65.83\%           & -\\  
& DeMamba-XCLIP-PT \cite{chen2024demamba} & 79.31\%           & - \\  
& XCLIP-B-FT \cite{chen2024demamba}       & 86.07\%           & - \\ \arrayrulecolor[gray]{0.85} \cline{2-4} \arrayrulecolor{black} 
& \textbf{\methodname{}}                  & \cellcolor{\red{}}{\textbf{90.49\%}}  & \cellcolor{\red{}}{\textbf{92.67\%}} \\ \hline
\end{tabular}
}
\label{tab:sota_comparison_detection}
\end{table}
\section{Detection Performance}\label{suppl:detection_performance}
On recent and popularly used datasets (both face-manipulated and fully synthetic), \methodname{}'s DeepFake detection capabilities have been tested in Table \ref{tab:sota_comparison_detection}. Since these datasets lack text annotations for the anomalies, the reasoning metrics for these datasets cannot be computed and thus only detection performance have been reported. Note that for DF40 \cite{yan2024df40} and DeMamba \cite{chen2024demamba} datasets, all the competitor methods have been trained and tested on the same datasets (in-domain evaluation). However, \methodname{} has been tested in purely cross-domain setting, having been trained on DD-VQA \cite{zhang2024common} and DVF \cite{song2024learning} still outperforms the state-of-the-art methods in detection performance.

The DeMamba results also provide a strong control for potential data-leakage concerns on our own DVF split. \methodname{} is trained on DVF and evaluated on DeMamba, an entirely different set of generators, and still outperforms competitors evaluated in-domain on DeMamba itself. Performance on a benchmark unseen during training cannot be explained by any within-dataset leakage, which further reinforces the generalization claims of Sec.~\ref{sec:experiments}.

\begin{table}[]
\centering
\caption{\textbf{Ablation on data compression factors:} Results obtained by the \methodname{} model on different compression factors of the FF++ \cite{rossler2019faceforensics++} dataset using the same language annotations from DD-VQA \cite{zhang2024common}. The small differences across the compression factors highlight the robustness of \methodname{}.}
\vspace{-1em}
\resizebox{\columnwidth}{!}{
\begin{tabular}{c|c|c|c|c|c}
\hline
\textbf{Compression} & \textbf{Accuracy} $\uparrow$ & \textbf{BLEU-3} $\uparrow$ & \textbf{BLUE\_4} $\uparrow$ & \textbf{ROUGE-L}  $\uparrow$ & \textbf{CIDEr} $\uparrow$ \\ \hline
raw                  & 94.17\%           & 0.4587           & 0.4131           & 0.6270            & 2.6109         \\
c23                  & 93.61\%           & 0.4610           & 0.4269           & 0.6103            & 2.6057         \\
c40                  & 94.12\%           & 0.4649           & 0.4304           & 0.6285            & 2.6321         \\ \hline
\end{tabular}
}
\vspace{-1em}
\label{tab:ablation_resolutions}
\end{table}
\section{Ablation on Resolutions}\label{suppl:resolutions}
We evaluate the robustness of \methodname{} to varying levels of image compression using the FF++ dataset \cite{rossler2019faceforensics++} benchmark, considering the ``raw", ``c23", and ``c40" compression settings commonly used in forensic literature. For this analysis, the language annotations (and data splits) are adopted from DD-VQA \cite{zhang2024common} to provide a consistent evaluation protocol across all settings.

As summarized in Table \ref{tab:ablation_resolutions}, \methodname{} achieves stable performance across all compression levels, with only minimal fluctuations in detection accuracy and explanation metrics (BLEU-3, BLEU-4, ROUGE-L, CIDEr). Notably, the accuracy remains above 93.6\% in all cases, and the scores on the language-based explainability metrics (e.g., BLEU-4 and CIDEr) remain consistently high, despite the potential visual degradation introduced by aggressive compression (e.g., ``c40''). These results demonstrate that \methodname{} is highly robust to changes in input image quality, maintaining both detection effectiveness and the detailed quality of its region-grounded explanations, even under unfavorable, low-fidelity conditions.

This resilience is particularly important in practical forensic workflows, where manipulated content is often subject to recompression or quality loss due to online sharing or platform-imposed constraints. By showing reliable performance across this spectrum, \methodname{} provides consistent and trustworthy DeepFake analysis regardless of compression artifacts or image degradation.

\begin{figure*}
    \centering
    \includegraphics[width=0.99\textwidth]{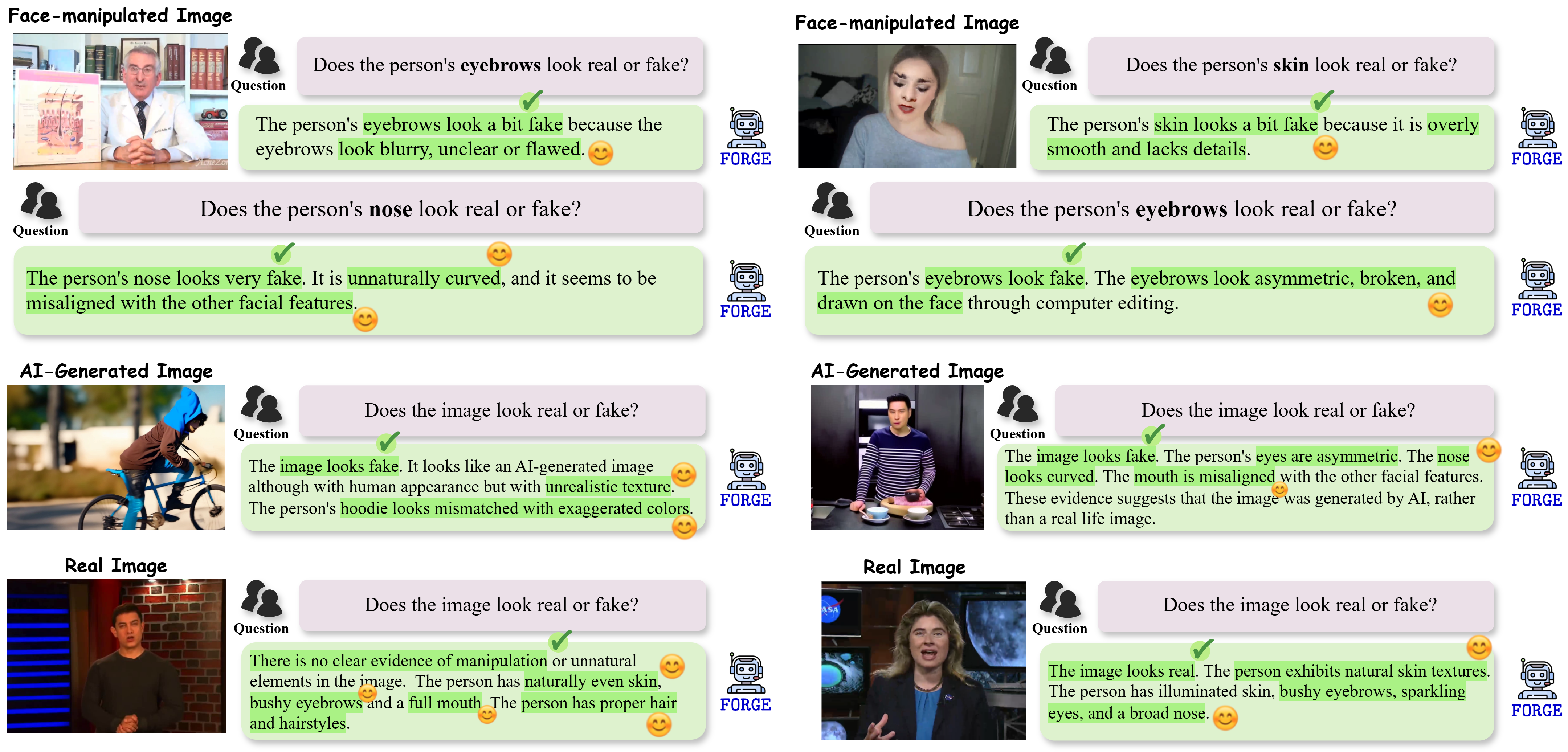}
    \vspace{-1em}
    \caption{\textbf{Qualitative examples} demonstrating \methodname{}'s detailed, region-grounded explanations across diverse image types. For face-manipulated images, \methodname{} identifies subtle inconsistencies such as blurry or flawed eyebrows, unnatural nose shapes, and misaligned facial features. In AI-generated images, it captures both global and local anomalies, including unrealistic textures, mismatched colors, and asymmetric facial components. For authentic images, the model confirms natural, harmonious features without producing false positives. Each example illustrates \methodname{}'s capability to answer fine-grained, attribute-specific queries, providing forensic insights that enhance transparency and interpretability well beyond traditional binary classification methods.}
    \label{fig:suppl_qualitative}
\end{figure*}

\section{More Qualitative Examples}\label{suppl:qualitative}
To further illustrate the effectiveness of \methodname{} in generating precise, interpretable explanations, we present additional qualitative results on a variety of image types in Figure \ref{fig:suppl_qualitative}, including face-manipulated DeepFakes, fully AI-generated images, and authentic real-world images. These examples were selected to demonstrate both the breadth and subtlety of reasoning achieved by our method across manipulation scenarios.

\noindent
\textbf{Face-Manipulated Images:} In face-manipulated examples, \methodname{} demonstrates its core strength: linking model predictions to region-specific visual evidence. For instance, when asked “Does the person’s eyebrows look real or fake?”, \methodname{} does not merely provide a binary answer, but instead grounds its explanation in visual attributes, remarking that the “eyebrows look a bit fake because they appear blurry, unclear or flawed,” or “the eyebrows look fake, asymmetric, broken, and drawn on the face through computer editing.” Similarly, when queried about the nose, the model observes that “the person’s nose looks very fake, is unnaturally curved, and misaligned with the other facial features.” When asked about the skin, the response cites “overly smooth skin lacking details.” Such attribute-localized, textually explicit responses demonstrate not only detection accuracy, but also the interpretability and transparency required for practical DeepFake forensic workflows. By surfacing minute, manipulation-specific artifacts overlooked by conventional MLLMs, \methodname{} enables investigators to probe exactly where and how an image has been altered.

\noindent
\textbf{AI-Generated Images:} For fully AI-generated content, \methodname{} fuses holistic scene analysis with attribute-level reasoning. In these examples, when asked whether the image appears real or fake, the model explicitly identifies telltale cues of synthesis: “the image looks fake,” referencing “AI-generated appearance with unrealistic texture,” or observing that “the person’s hoodie looks mismatched with exaggerated colors.” In another case, the model details that “the image looks fake; the person’s eyes are asymmetric, the nose looks curved, and the mouth is misaligned with other facial features,” clearly indicating knowledge of how AI-generation often fails to capture coherent facial structure. These multi-faceted, region-specific explanations set \methodname{} apart from prior methods that offer only undifferentiated global judgements, and highlight the model’s utility for both forensic screening and downstream content auditing.

\noindent
\textbf{Real Images:} For authentic images, \methodname{} maintains high precision and does not falsely attribute benign artifacts as fake. Its answers highlight the absence of manipulation with clarifying details: acknowledging “naturally even skin, bushy eyebrows, and a full mouth,” or commenting that “the person has proper hair and hairstyles,” “illuminated skin, sparkling eyes, and a broad nose.” This restraint, coupled with accurate grounding, ensures trustworthiness and minimizes false alarms in real-world deployment scenarios.

\noindent
\textbf{Analysis and Implications:} These qualitative results underscore several key strengths of \methodname{}. First, the model’s ability to align its explanations with concrete, localized regions in the image ensures that each rationale is visually and forensically grounded. This explicit region-level correspondence builds trust in the model’s predictions and directly addresses the limitations of prior methods that rely solely on global analysis or binary outputs.

Second, \methodname{} demonstrates remarkable flexibility in handling both open-ended and attribute-specific forensic queries. Whether prompted about the overall authenticity of an image (“Does the image look real or fake?”) or about specific facial features (such as \texttt{eyebrows}, \texttt{nose}, or \texttt{skin}), \methodname{} provides granular, context-aware explanations. For face-manipulated images, the system surfaces manipulation artifacts like blurry or asymmetric eyebrows, unnaturally curved noses, or unnaturally smooth skin, clearly indicating both the presence and nature of DeepFake manipulations. For AI-generated content, \methodname{} is able to synthesize holistic scene cues with localized anomalies, referencing unnatural textures, mismatched colors, and misaligned facial features. Real images are correctly described with natural, harmonious feature attributions, showing the system’s precision and its resistance to false positives.

A particularly notable advantage, as evidenced across these examples, is \methodname{}’s competency in addressing fine-grained, query-driven prompts. By moving beyond simple authenticity judgments, and providing attribute-resolved rationales, the model empowers forensic experts with actionable insights, identifying precisely which regions or features are manipulated and explaining why.

Collectively, these qualitative results highlight \methodname{}’s unique contributions: robust region-level grounding, the ability to deliver interpretable and visually anchored rationales, and consistent performance across both fine-grained and holistic queries. By bridging the gap between detection accuracy and explanation transparency, \methodname{} advances the state of the art and offers a practical, reliable tool for explainable DeepFake forensics.

\begin{table}
\centering
\caption{Evaluation of \methodname{} stratified by finegrained region or scene aspect.}
\vspace{-1em}
\resizebox{\linewidth}{!}{
\begin{tabular}{l|l|c|c|c|c|c}
\hline
\textbf{Dataset} & \textbf{Region/Aspect} & \textbf{Accuracy} & \textbf{AUC} & \textbf{BLEU-4} & \textbf{ROUGE-L} & \textbf{CIDEr} \\
\hline
\multirow{6}{*}{\begin{tabular}[c]{@{}c@{}}DD-VQA\\ \cite{zhang2024common} \end{tabular}} 
 & Eyes              & 95.82 & 96.46 & 0.4521 & 0.6452 & 2.7813 \\
 & Nose              & 95.14 & 96.11 & 0.4463 & 0.6398 & 2.6954 \\
 & Mouth             & 94.61 & 95.63 & 0.4387 & 0.6331 & 2.6612 \\
 & Eyebrows          & 92.37 & 94.17 & 0.4015 & 0.6024 & 2.4217 \\
 & Skin              & 93.75 & 95.45 & 0.4249 & 0.6196 & 2.5839 \\
 \arrayrulecolor[gray]{0.85} \cline{2-6} \arrayrulecolor{black}
 & \textit{Overall}           & 94.12 & 95.39 & 0.4304 & 0.6285 & 2.6321 \\
\hline
\multirow{8}{*}{\begin{tabular}[c]{@{}c@{}}DVF\\ \cite{song2024learning}\end{tabular}} 
 & Background        & 95.35 & 96.67 & 0.4412 & 0.5487 & 2.2615 \\
 & Lighting          & 94.94 & 96.19 & 0.4375 & 0.5421 & 2.2328 \\
 & Shadows           & 94.37 & 95.71 & 0.4318 & 0.5364 & 2.2012 \\
 & Textures          & 93.89 & 95.23 & 0.4236 & 0.5279 & 2.1527 \\
 & Reflections       & 93.10 & 94.74 & 0.4182 & 0.5213 & 2.1049 \\
 & Object Boundaries & 94.12 & 95.43 & 0.4291 & 0.5339 & 2.1853 \\
 & Color Consistency & 93.57 & 95.01 & 0.4210 & 0.5260 & 2.1396 \\
 \arrayrulecolor[gray]{0.85} \cline{2-6} \arrayrulecolor{black}
 & \textit{Overall}           & 94.47 & 95.82 & 0.4279 & 0.5338 & 2.1744 \\
\hline
\end{tabular}
}
\label{tab:stratified}
\vspace{-2em}
\end{table}
\section{Stratified Evaluation by Region and Aspect}\label{suppl:stratified}
The region-referential evaluation reports aggregate faithfulness and hallucination metrics. In Table~\ref{tab:stratified} we report \methodname{}'s performance broken down by the specific facial region (DD-VQA) or scene aspect (DVF) that each query targets, providing a finer-grained view of where the model is strong and where residual error concentrates. Table~\ref{tab:stratified} reports accuracy, AUC, and the three generation metrics per region and aspect.

On DD-VQA, performance is highest on the regions that carry the most discriminative manipulation evidence: eyes (95.82\% accuracy) and nose (95.14\%), both of which exhibit pronounced geometric and blending artifacts under face manipulation. Performance is lowest on eyebrows (92.37\%), consistent with their smaller spatial extent and the subtlety of eyebrow-level edits. On DVF, the model is strongest on background and lighting (95.35\% and 94.94\%), where synthesis artifacts are spatially broad, and weakest on reflections (93.10\%) and color consistency (93.57\%), which require global cross-region comparison and are the aspects most often ambiguous even to human annotators. Across both datasets the spread between the best and worst region is under 3.5 accuracy points, indicating that the dense-patch stream surfaces evidence consistently rather than concentrating performance on a single easy region.

\end{document}